\documentclass[nohyperref]{article}

\usepackage{microtype}
\usepackage{graphicx}
\usepackage{subfigure}
\usepackage{bm}
\usepackage{booktabs} 

\usepackage{hyperref}

\usepackage{caption}

\usepackage[accepted]{setup/icml2022}

\usepackage{xcolor}

\usepackage{amsmath,amsfonts,amsthm,amssymb}
\usepackage{mathtools}

\DeclareMathOperator*{\argmax}{arg\,max}
\DeclareMathOperator*{\argmin}{arg\,min}

\usepackage{babel}
\usepackage[title]{appendix}
\usepackage{titlesec}
\usepackage{tikz}
\usetikzlibrary{shapes.geometric}
\usetikzlibrary{arrows.meta,arrows}
\usepackage{tabularx}
\usepackage{stackengine}
\usepackage{xspace}
\usepackage{stfloats}
\usepackage[hang,flushmargin]{footmisc}

\definecolor{cred}{HTML}{D62728}
\definecolor{cblue}{HTML}{1F77B4}
\definecolor{cgreen}{HTML}{79AB76}
\definecolor{cgrey}{rgb}{0.6,0.6,0.6}

\definecolor{highlight}{rgb}{0,0,0}

\newcommand{\st}{\mathbf{s}_t}
\newcommand{\stp}{\mathbf{s}_{t+1}}

\newcommand{\at}{\mathbf{a}_t}

\newcommand{\bs}{\mathbf{s}}
\newcommand{\ba}{\mathbf{a}}

\newcommand{\btau}[1]{\bm{\tau}^{#1}}
\newcommand{\opt}{\mathcal{O}}

\newcommand{\expect}[2]{\mathbb{E}_{#1} \left[ #2 \right] }

\newcommand{\te}[1]{\texttt{#1}}

\def\vf{{\bm{f}}}

\def\vs{{\bm{s}}}

\newcommand{\highlight}[1]{\textbf{#1}}

\newcommand\blfootnote[1]{%
  \begingroup
  \renewcommand\thefootnote{}\footnote{#1}%
  \addtocounter{footnote}{-1}%
  \endgroup
}

\newcommand{\fig}[1]{Figure~\ref{#1}}

\newcommand{\myparagraph}[1]{{\bf #1}\quad}

\DeclarePairedDelimiterX{\infdivx}[2]{(}{)}{%
  #1\;\delimsize|\delimsize|\;#2%
}

\newcommand{\method}{Diffuser\xspace}

\icmltitlerunning{Planning with Diffusion for Flexible Behavior Synthesis}

\begin{document}

\twocolumn[
\icmltitle{Planning with Diffusion for Flexible Behavior Synthesis}

\icmlsetsymbol{equal}{*}

\begin{icmlauthorlist}
\icmlauthor{Michael Janner}{equal,berkeley}
\icmlauthor{Yilun Du}{equal,mit}
\icmlauthor{Joshua B. Tenenbaum}{mit}
\icmlauthor{Sergey Levine}{berkeley}
\end{icmlauthorlist}

\icmlaffiliation{berkeley}{University of California, Berkeley}
\icmlaffiliation{mit}{MIT}

\icmlcorrespondingauthor{}{janner@berkeley.edu}
\icmlcorrespondingauthor{}{yilundu@mit.edu}

\icmlkeywords{Planning, deep reinforcement learning, diffusion models, denoising}

\vskip 0.3in]

\printAffiliationsAndNotice{\icmlEqualContribution} 

\begin{abstract}
Model-based reinforcement learning methods often use learning only for the purpose of estimating an approximate dynamics model, offloading the rest of the decision-making work to classical trajectory optimizers.
While conceptually simple, this combination has a number of empirical shortcomings, suggesting that learned models may not be well-suited to standard trajectory optimization.
In this paper, we consider what it would look like to fold as much of the trajectory optimization pipeline as possible into the modeling problem, such that sampling from the model and planning with it become nearly identical.
The core of our technical approach lies in a diffusion probabilistic model that plans by iteratively denoising trajectories.
We show how classifier-guided sampling and image inpainting can be reinterpreted as coherent planning strategies, explore the unusual and useful properties of diffusion-based planning methods, and demonstrate the effectiveness of our framework in control settings that emphasize long-horizon decision-making and test-time flexibility.
\end{abstract}

\section{Introduction}
\label{sec:intro}

Planning with a learned model is a conceptually simple framework for reinforcement learning and data-driven decision-making.
Its appeal comes from employing learning techniques only where they are the most mature and effective: for the approximation of unknown environment dynamics in what amounts to a supervised learning problem.
Afterwards, the learned model may be plugged into classical trajectory optimization routines \cite{tassa2012synthesis,posa2014direct,matthew2017trajectory},
which are similarly well-understood in their original context.
\vspace{1em}

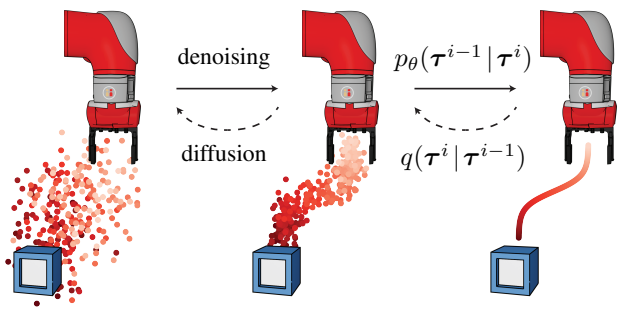
\begin{figure}
\centering
\begin{tikzpicture}
    \node[anchor=south west,inner sep=0] (image) at (0,0)
        {\includegraphics[width=\columnwidth]{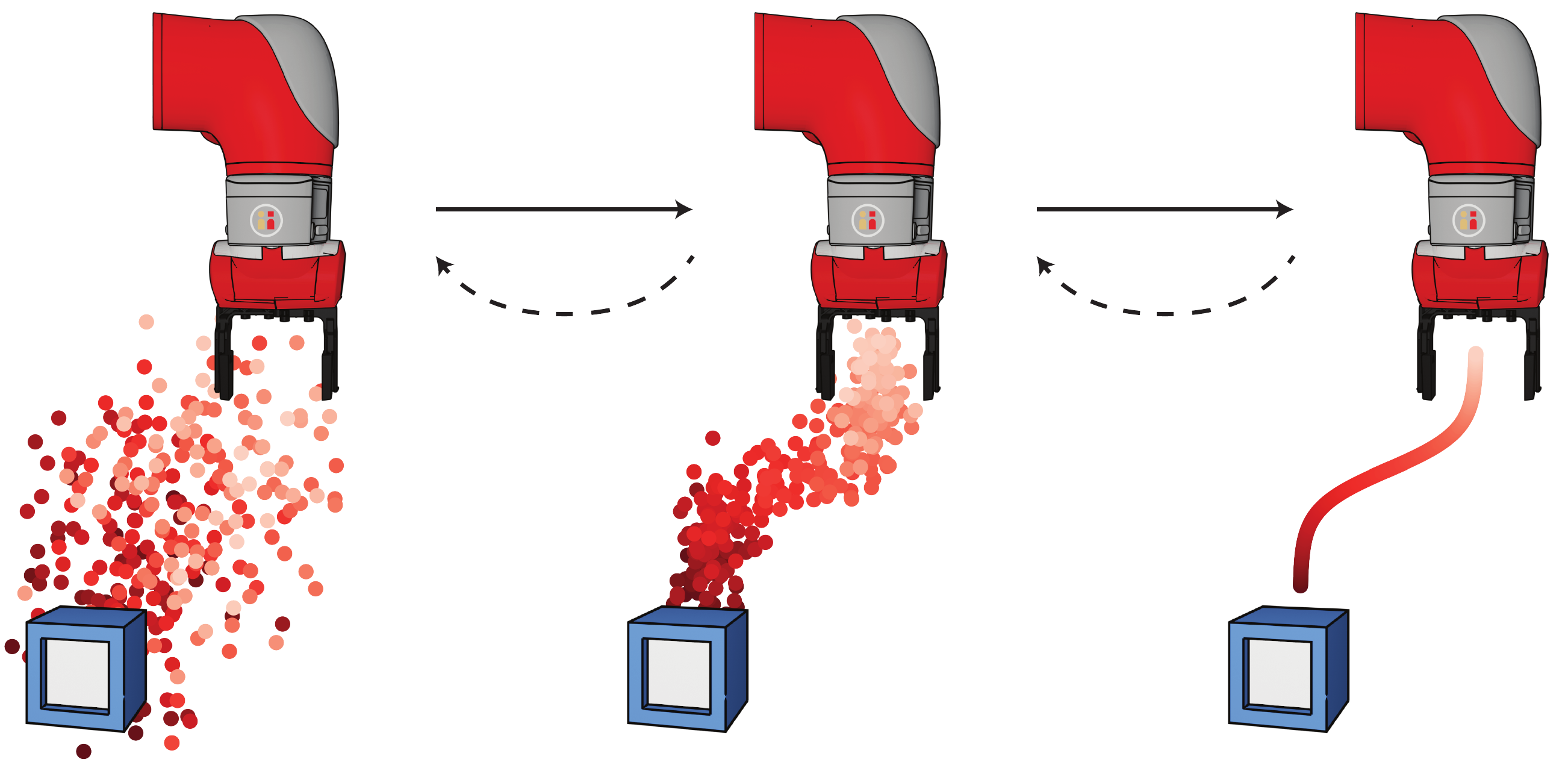}};
    \begin{scope}[x={(image.south east)},y={(image.north west)}]
        \node[] at (.74,.825) {$p_\theta(\btau{i-1} \!\mid\! \btau{i})$};
        \node[] at (.74,.5) {$q(\btau{i} \!\mid\! \btau{i-1})$};
        \node[] at (.355,.825) {\small denoising};
        \node[] at (.355,.515) {\small diffusion};
    \end{scope}
\end{tikzpicture}
\vspace{-1.5em}
\caption{
    \method{} is a diffusion probabilistic model that plans by iteratively refining trajectories.
}
\vspace{-1em}
\label{fig:teaser}
\end{figure}

However, this combination rarely works as described.
Because powerful trajectory optimizers exploit learned models, plans generated by this procedure often look more like adversarial examples than optimal trajectories \cite{talvitie2014hallcuinated,ke2018modeling}.
As a result, contemporary model-based reinforcement learning algorithms often inherit more from model-free methods, such as value functions and policy gradients \cite{wang2019benchmarking}, than from the trajectory optimization toolbox.
Those methods that do rely on online planning tend to use simple gradient-free trajectory optimization routines like random shooting \cite{nagabandi2018mbmf} or the cross-entropy method \cite{botev2013cem,chua2018pets} to avoid the aforementioned issues.
\blfootnote{
\noindent Code and visualizations of the learned denoising process are available at \textbf{\href{https://diffusion-planning.github.io/}
    {\texttt{diffusion-planning.github.io}}}.
}

In this work, we propose an alternative approach to data-driven trajectory optimization.
The core idea is to train a model that is directly amenable to trajectory optimization, in the sense that sampling from the model and planning with it become nearly identical.
This goal requires a shift in how the model is designed.
Because learned dynamics models are normally meant to be proxies
for environment dynamics, improvements are often achieved by structuring the model according to the underlying causal process \cite{bapst2019structured}.
Instead, we consider how to design a model in line with the planning problem in which it will be used.
For example, because the model will ultimately be used for planning, action distributions are just as important as state dynamics and long-horizon accuracy is more important than single-step error.
On the other hand, the model should remain agnostic to reward function so that it may be used in multiple tasks, including those unseen during training.
Finally, the model should be designed so that its plans, and not just its predictions, improve with experience and are resistant to the myopic failure modes of standard shooting-based planning algorithms.

\begin{figure}
\centering
\includegraphics[width=\columnwidth]{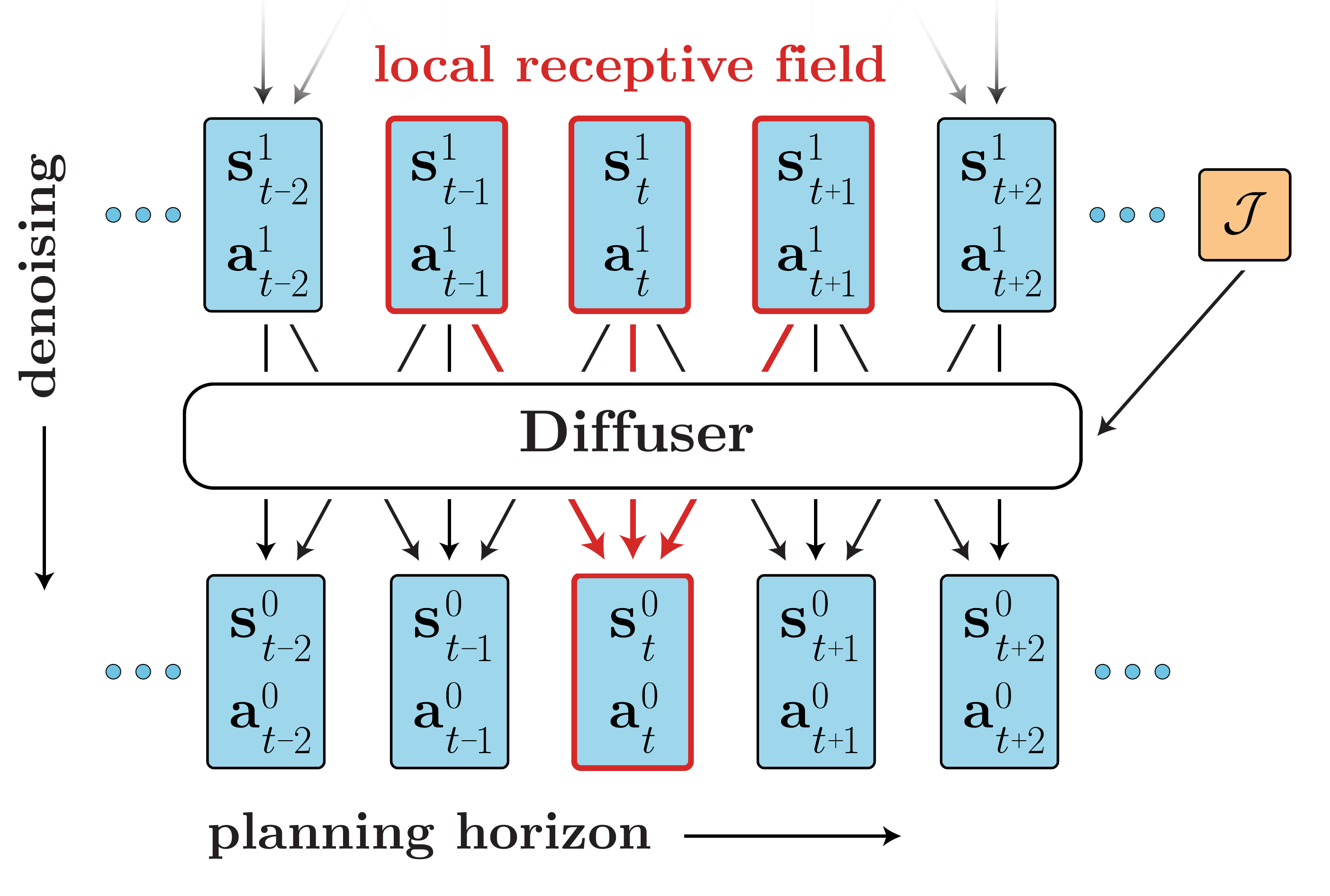}
\caption{
    Diffuser samples plans by iteratively denoising two-dimensional arrays consisting of a variable number of state-action pairs.
    A small receptive field constrains the model to only enforce local consistency during a single denoising step.
    By composing many denoising steps together, local consistency can drive global coherence of a sampled plan.
    An optional guide function $\mathcal{J}$ can be used to bias plans toward those optimizing a test-time objective or satisfying a set of constraints.
}
\vspace{-1.5em}
\label{fig:architecture}
\end{figure}

We instantiate this idea as a trajectory-level diffusion probabilistic model \citep{sohldickstein2015nonequilibrium,ho2020denoising} called \method{}, visualized in Figure~\ref{fig:architecture}.
Whereas standard model-based planning techniques predict forward in time autoregressively, \method{} predicts all timesteps of a plan simultaneously.
The iterative sampling process of diffusion models leads to flexible conditioning, allowing for auxiliary guides to modify the sampling procedure to recover trajectories with high return or satisfying a set of constraints.
This formulation of data-driven trajectory optimization has several appealing properties:

\textbf{Long-horizon scalability~~}
Diffuser is trained for the accuracy of its generated trajectories rather than its single-step error, so it does not suffer from the compounding rollout errors of single-step dynamics models and scales more gracefully with respect to long planning horizon.

\textbf{Task compositionality~~}
Reward functions provide auxiliary gradients to be used while sampling a plan, allowing for a straightforward way of planning by composing multiple rewards simultaneously by adding together their gradients.

\textbf{Temporal compositionality~~}
Diffuser generates globally coherent trajectories by iteratively improving local consistency, allowing it to generalize to novel trajectories by stitching together in-distribution subsequences.

\textbf{Effective non-greedy planning~~}
By blurring the line between model and planner, the training procedure that improves the model's predictions also has the effect of improving its planning capabilities.
This design yields a learned planner that can solve the types of long-horizon, sparse-reward problems that prove difficult for many conventional planning methods.

The core contribution of this work is a denoising diffusion model designed for trajectory data and an associated probabilistic framework for behavior synthesis.
While unconventional compared to the types of models routinely used in deep model-based reinforcement learning, we demonstrate that \method has a number of useful properties and is particularly effective in offline control settings that require long-horizon reasoning and test-time flexibility.

\section{Background}
\label{sec:background}

Our approach to planning is a learning-based analogue of past work in behavioral synthesis using trajectory optimization \citep{witkin1988spacetime, tassa2012synthesis}.
In this section, we provide a brief background on the problem setting considered by trajectory optimization and the class of generative models we employ for that problem.

\subsection{Problem Setting}
\label{sec:background_traj_opt}
Consider a system governed by the discrete-time dynamics $\stp = \vf(\st, \at)$ at state $\st$ given an action $\at$.
Trajectory optimization refers to finding a sequence of actions $\ba_{0:T}^*$ that maximizes (or minimizes) an objective $\mathcal{J}$ factorized over per-timestep rewards (or costs) $r(\st, \at)$:
\[
\ba_{0:T}^* = \argmax_{\ba_{0:T}} \mathcal{J}(\bs_0, \ba_{0:T}) = \argmax_{\ba_{0:T}} \sum_{t=0}^{T} r(\st, \at)
\]
where $T$ is the planning horizon.
We use the abbreviation $\btau{} = (\bs_0, \ba_0, \bs_1, \ba_1, \ldots, \bs_T, \ba_T)$ to refer to a trajectory of interleaved states and actions and $\mathcal{J}(\btau{})$ to denote the objective value of that trajectory.

\subsection{Diffusion Probabilistic Models}
\label{sec:background_diffusion}
Diffusion probabilistic models \citep{sohldickstein2015nonequilibrium,ho2020denoising} pose the data-generating process as an iterative denoising procedure $p_\theta(\btau{i-1} \mid \btau{i})$.
This denoising is the reverse of a forward diffusion process $q(\btau{i} \mid \btau{i-1})$ that slowly corrupts the structure in data by adding noise.
The data distribution induced by the model is given by:
\[
p_\theta(\btau{0}) =
\int p(\btau{N}) \prod_{i=1}^{N} p_\theta(\btau{i-1} \mid \btau{i}) \mathrm{d} \btau{1:N}
\]
where $p(\btau{N})$ is a standard Gaussian prior and $\btau{0}$ denotes (noiseless) data.
Parameters $\theta$ are optimized by minimizing a variational bound on the negative log likelihood of the reverse process:
$
\theta^* = \argmin_{\theta}
-\expect{\btau{0}}{\log p_\theta(\btau{0})}.
$
The reverse process is often parameterized as Gaussian with fixed timestep-dependent covariances:
\[
p_\theta(\btau{i-1} \mid \btau{i}) = \mathcal{N}(\btau{i-1} \mid \mu_\theta(\btau{i}, i), \Sigma^i).
\]
The forward process $q(\btau{i} \mid \btau{i-1})$ is typically prespecified.

\textbf{Notation.}~~
There are two ``times" at play in this work: that of the diffusion process and that of the planning problem.
We use superscripts ($i$ when unspecified) to denote diffusion timestep and subscripts ($t$ when unspecified) to denote planning timestep.
For example, $\st^{0}$ refers to the $t^{\text{th}}$ state in a noiseless trajectory.
When it is unambiguous from context, superscripts of noiseless quantities are omitted: $\btau{} = \btau{0}$.
We overload notation slightly by referring to the $t^\text{th}$ state (or action) in a trajectory $\btau{}$ as $\btau{}_{\st}$ (or $\btau{}_{\at}$).

\section{Planning with Diffusion}
\label{sec:method}

A major obstacle to using trajectory optimization techniques is that they require knowledge of the environment dynamics $\vf$.
Most learning-based methods attempt to overcome this obstacle by training an approximate dynamics model and plugging it in to a conventional planning routine.
However, learned models are often poorly suited to the types of planning algorithms designed with ground-truth models in mind, leading to planners that exploit learned models by finding adversarial examples.

We propose a tighter coupling between modeling and planning.
Instead of using a learned model in the context of a classical planner, we subsume as much of the planning process as possible into the generative modeling framework, such that planning becomes nearly identical to sampling.
We do this using a diffusion model of trajectories, $p_\theta(\btau{})$.
The iterative denoising process of a diffusion model lends itself to flexible conditioning by way of sampling from perturbed distributions of the form:
\begin{equation}
\label{eq:perturbed}
\tilde{p}_\theta(\btau{}) \propto p_\theta(\btau{}) h(\btau{}).
\end{equation}

The function $h(\btau{})$ can contain information about prior evidence (such as an observation history), desired outcomes (such as a goal to reach), or general functions to optimize (such as rewards or costs).
Performing inference in this perturbed distribution can be seen as a probabilistic analogue to the trajectory optimization problem posed in Section~\ref{sec:background_traj_opt}, as it requires finding trajectories that are both physically realistic under $p_\theta(\btau{})$ and high-reward (or constraint-satisfying) under $h(\btau{})$.
Because the dynamics information is separated from the perturbation distribution $h(\btau{})$, a single diffusion model $p_\theta(\btau{})$ may be reused for multiple tasks in the same environment.

In this section, we describe \method, a diffusion model designed for learned trajectory optimization.
We then discuss two specific instantiations of planning with \method, realized as reinforcement learning counterparts to classifier-guided sampling and image inpainting.

\subsection{A Generative Model for Trajectory Planning}
\label{sec:model}

\textbf{Temporal ordering.}~~
Blurring the line between sampling from a trajectory model and planning with it yields an unusual constraint: we can no longer predict states autoregressively in temporal order.
Consider the goal-conditioned inference ${p(\bs_1 \mid \bs_0, \bs_T)}$;
the next state $\bs_1$ depends on a \emph{future} state as well as a prior one.
This example is an instance of a more general principle:
while dynamics prediction is causal, in the sense that the present is determined by the past, decision-making and control can be anti-causal, in the sense that decisions in the present are conditional on the future.\footnote{
    In general reinforcement learning contexts, conditioning on the future emerges from the assumption of future optimality for the purpose of writing a dynamic programming recursion. Concretely, this appears as the future optimality variables $\mathcal{O}_{t:T}$ in the action distribution $\log p(\at \mid \st, \mathcal{O}_{t:T})$ (Levine, 2018)\nocite{levine2018reinforcement}.
}
Because we cannot use a temporal autoregressive ordering, we design \method to predict all timesteps of a plan concurrently.

\textbf{Temporal locality.}~~
Despite not being autoregressive or Markovian, \method features a relaxed form of temporal locality.
In Figure~\ref{fig:architecture}, we depict a dependency graph for a diffusion model consisting of a single temporal convolution.
The receptive field of a given prediction only consists of nearby timesteps, both in the past and the future.
As a result, each step of the denoising process can only make predictions based on local consistency of the trajectory.
By composing many of these denoising steps together, however, local consistency can drive global coherence.

\textbf{Trajectory representation.}~~
\method is a model of trajectories designed for planning, meaning that the effectiveness of the controller derived from the model is just as important as the quality of the state predictions.
As a result, states and actions in a trajectory are predicted jointly; for the purposes of prediction the actions are simply additional dimensions of the state.
Specifically, we represent inputs (and outputs) of \method as a two-dimensional array:
\vspace{-.1cm}
\begin{align}
\label{eq:trajectory_array}
\btau{} = \begin{bmatrix}
    \bs_{0} & \bs_{1} & \raisebox{-.2cm}{\ldots} & \bs_{T} \\
    \ba_{0} & \ba_{1} & & \ba_{T}
\end{bmatrix}.
\end{align}
with one column per timestep of the planning horizon.

\textbf{Architecture.}~~
We now have the ingredients needed to specify a \method architecture:
\textbf{(1)} an entire trajectory should be predicted non-autoregressively,
\textbf{(2)} each step of the denoising process should be temporally local, and
\textbf{(3)} the trajectory representation should allow for equivariance  along one dimension (the planning horizon) but not the other (the state and action features).
We satisfy these criteria with a model consisting of repeated (temporal) convolutional residual blocks.
The overall architecture resembles the types of U-Nets that have found success in image-based diffusion models, but with two-dimensional spatial convolutions replaced by one-dimensional temporal convolutions (Figure~\ref{fig:app_architecture}).
Because the model is fully convolutional, the horizon of the predictions is determined not by the model architecture, but by the input dimensionality; it can change dynamically during planning if desired.

\textbf{Training.}~~
We use \method to parameterize a learned gradient $\epsilon_\theta(\btau{i}, i)$ of the trajectory denoising process, from which the mean $\mu_\theta$ can be solved in closed form \citep{ho2020denoising}.
We use the simplified objective for training the $\epsilon$-model, given by:
\[
\mathcal{L}(\theta) = \expect{
    i,\epsilon,\btau{0}
    }{
    \lVert \epsilon - \epsilon_\theta(\btau{i}, i) \rVert^2
    },
\]
in which $i \sim \mathcal{U}\{1, 2, \ldots, N\}$ is the diffusion timestep, $\epsilon \sim \mathcal{N}(\bm{0}, \bm{I})$ is the noise target, and $\btau{i}$ is the trajectory $\btau{0}$ corrupted with noise $\epsilon$.
Reverse process covariances $\Sigma^i$ follow the cosine schedule of \citet{nichol2021improved}.

\def\NoNumber#1{{\def\alglinenumber##1{}\STATE #1}\addtocounter{ALG@line}{-1}}

{\centering
\begin{figure}[t]
\begin{minipage}{\linewidth}
  \begin{algorithm}[H]
    \caption{Guided Diffusion Planning}
    \label{alg:rl}
    \begin{algorithmic}[1]
    \STATE \textbf{Require} Diffuser $\mu_\theta$, guide $\mathcal{J}$, scale $\alpha$, covariances $\Sigma^i$ \\
    \WHILE{not done}
        \STATE Observe state $\bs$; initialize plan  $\btau{N} \sim \mathcal{N}(\bm{0}, \bm{I})$ \\
        \FOR{$i = N, \ldots, 1$}
            \STATE \small{\color{gray}\te{// parameters of reverse transition}} \\
            \STATE $\mu \gets \mu_\theta(\btau{i})$ \\
            \STATE \small{\color{gray}\te{// guide using gradients of return}} 
            \STATE $\btau{i-1} \sim \mathcal{N}(\mu + \alpha \Sigma \nabla \mathcal{J}(\mu), \Sigma^i)$\\
            \STATE \small{\color{gray}\te{// constrain first state of plan}}
            \STATE $\btau{i-1}_{\vs_0} \gets \vs$ \hspace{0.15cm} \\
        \ENDFOR
        \STATE Execute first action of plan $\btau{0}_{\ba_0}$
    \ENDWHILE
    \end{algorithmic}
  \end{algorithm}
\end{minipage}
\vspace{-.2cm}
\end{figure}
}

\newcommand{\undersetimage}[3]{
    \begin{tikzpicture}
        \node[anchor=south west,inner sep=0] (image) at (0, 0)
            {\includegraphics[width=0.65cm]{#1}};
        \begin{scope}[x={(image.south east)},y={(image.north west)}]
            \node[] at (0.5, #3) {#2};
        \end{scope}
    \end{tikzpicture}
}

\begin{figure*}
\centering
\begin{minipage}[t]{0.55\textwidth}
    \raisebox{2cm}{\textbf{a}}~~~
    \hspace{.2cm}
    \includegraphics[width=2.5cm]{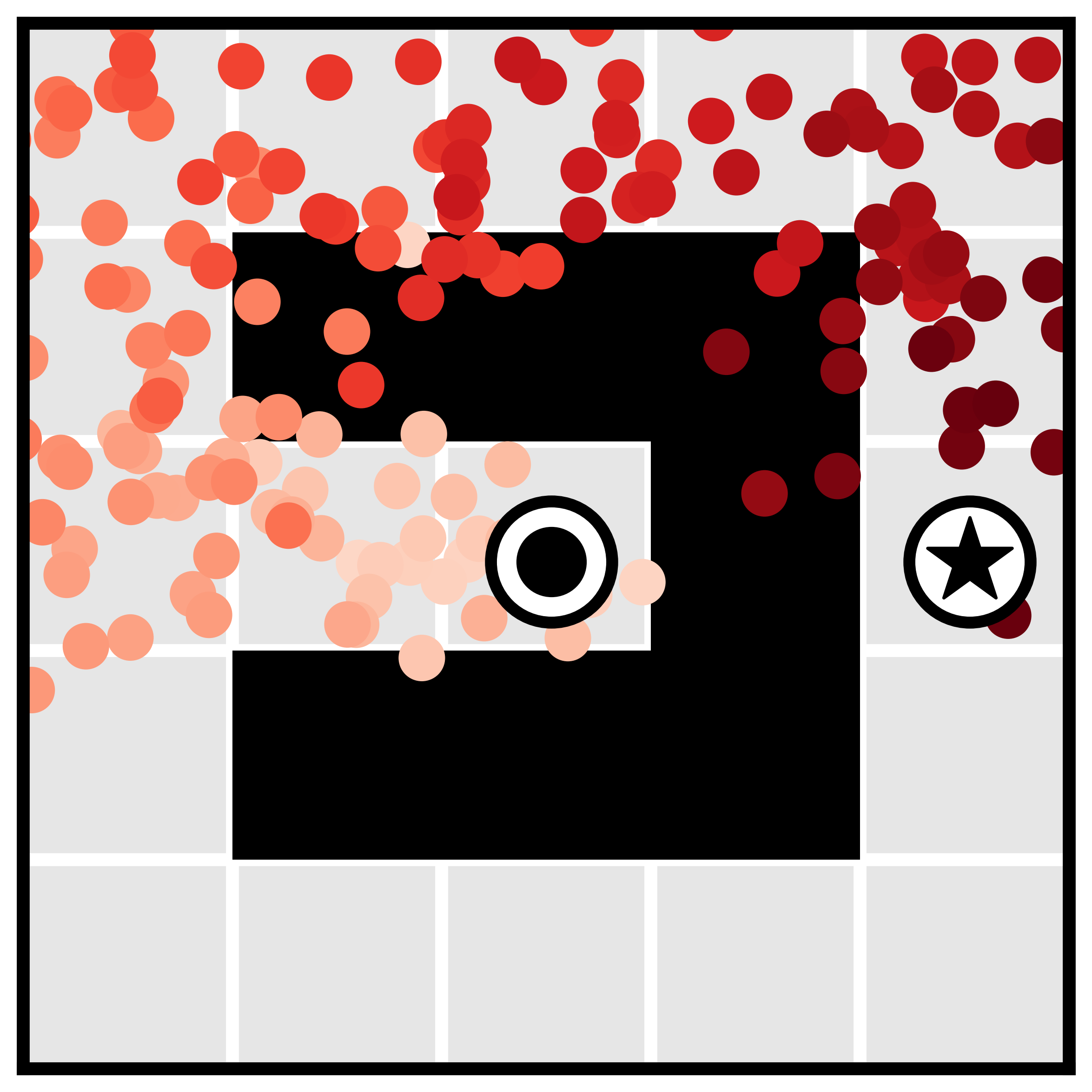}~
    \includegraphics[width=2.5cm]{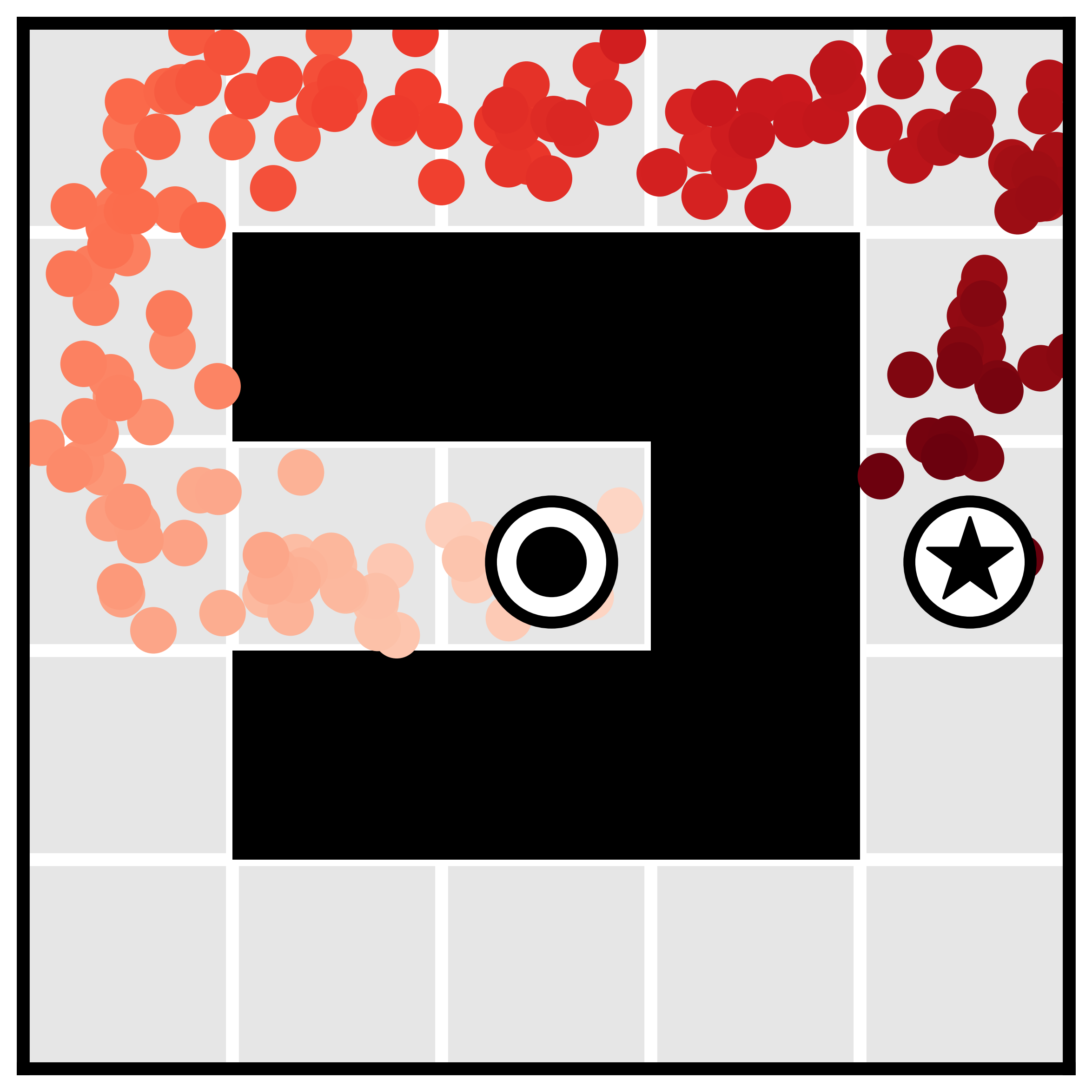}~
    \includegraphics[width=2.5cm]{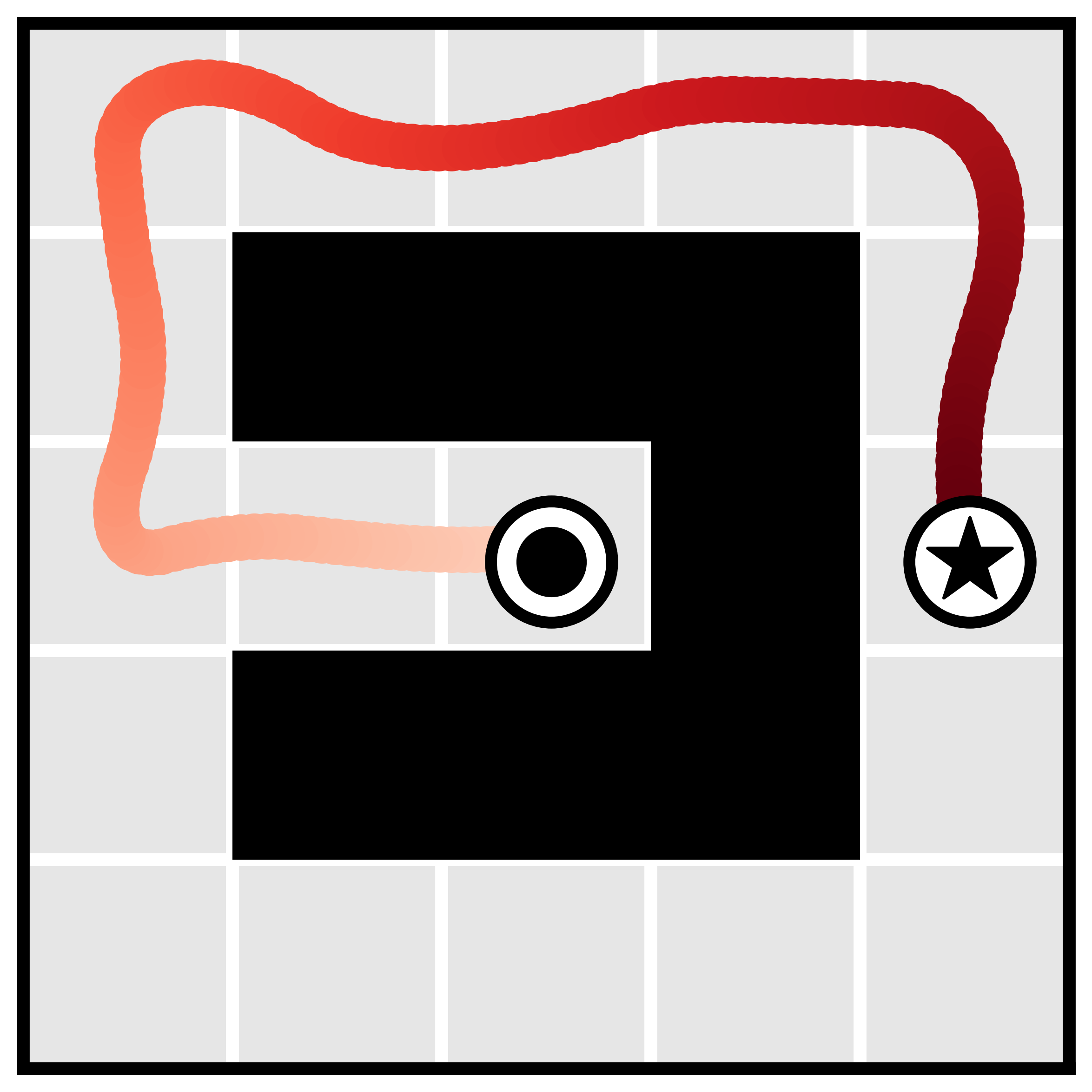} \\
    \vspace{-.4cm}
    \begin{center}
        \raisebox{.3\height}{denoising}~~
        \tikz \draw [-{Computer Modern Rightarrow[length=1.33mm, width=2mm]}, line width=.2mm] (0,0) -- (3,0); \\
    \end{center}
\end{minipage}%
\begin{minipage}[t]{0.35\textwidth}
    \raisebox{2cm}{\textbf{b}}~~~
    \includegraphics[width=2.5cm]{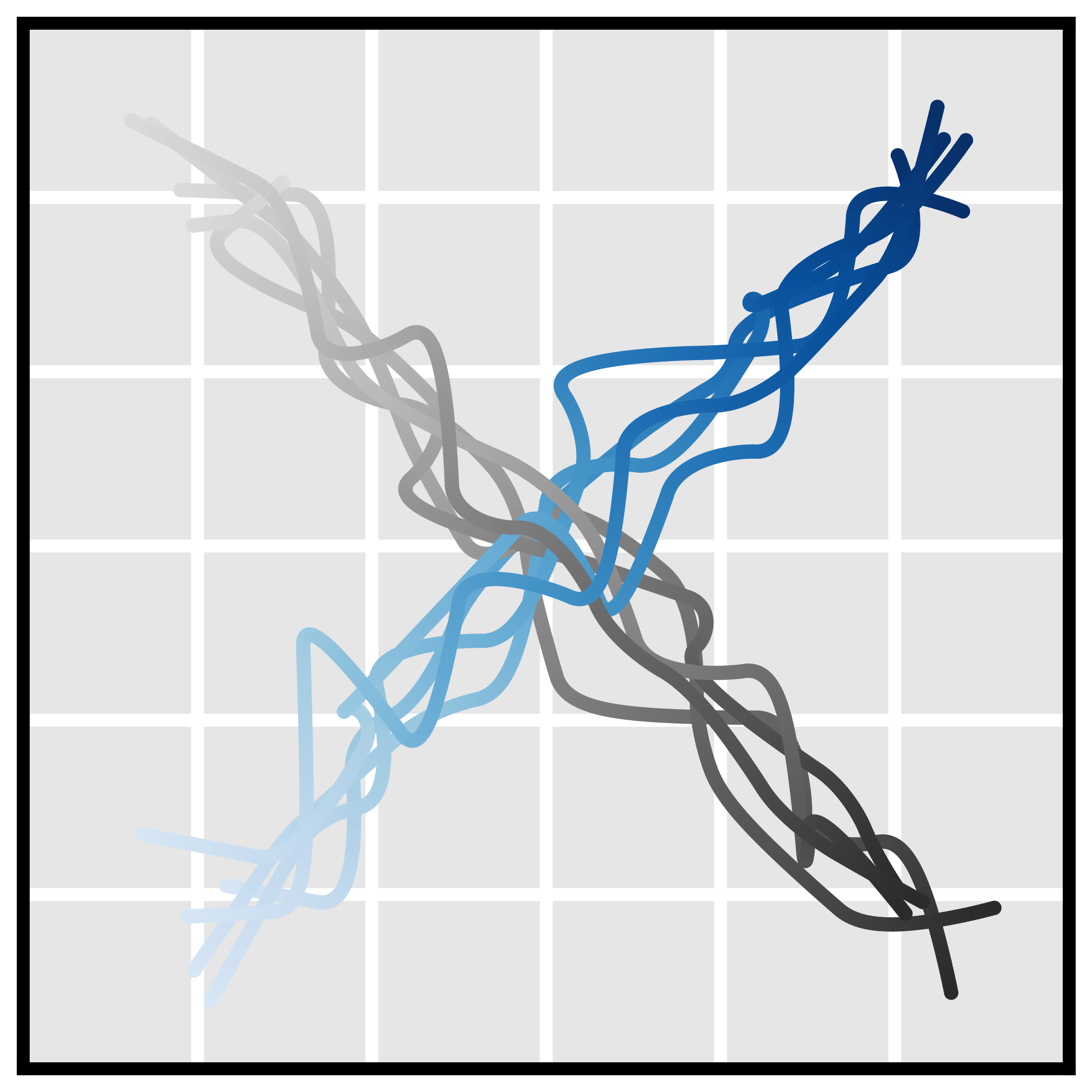}~
    \includegraphics[width=2.5cm]{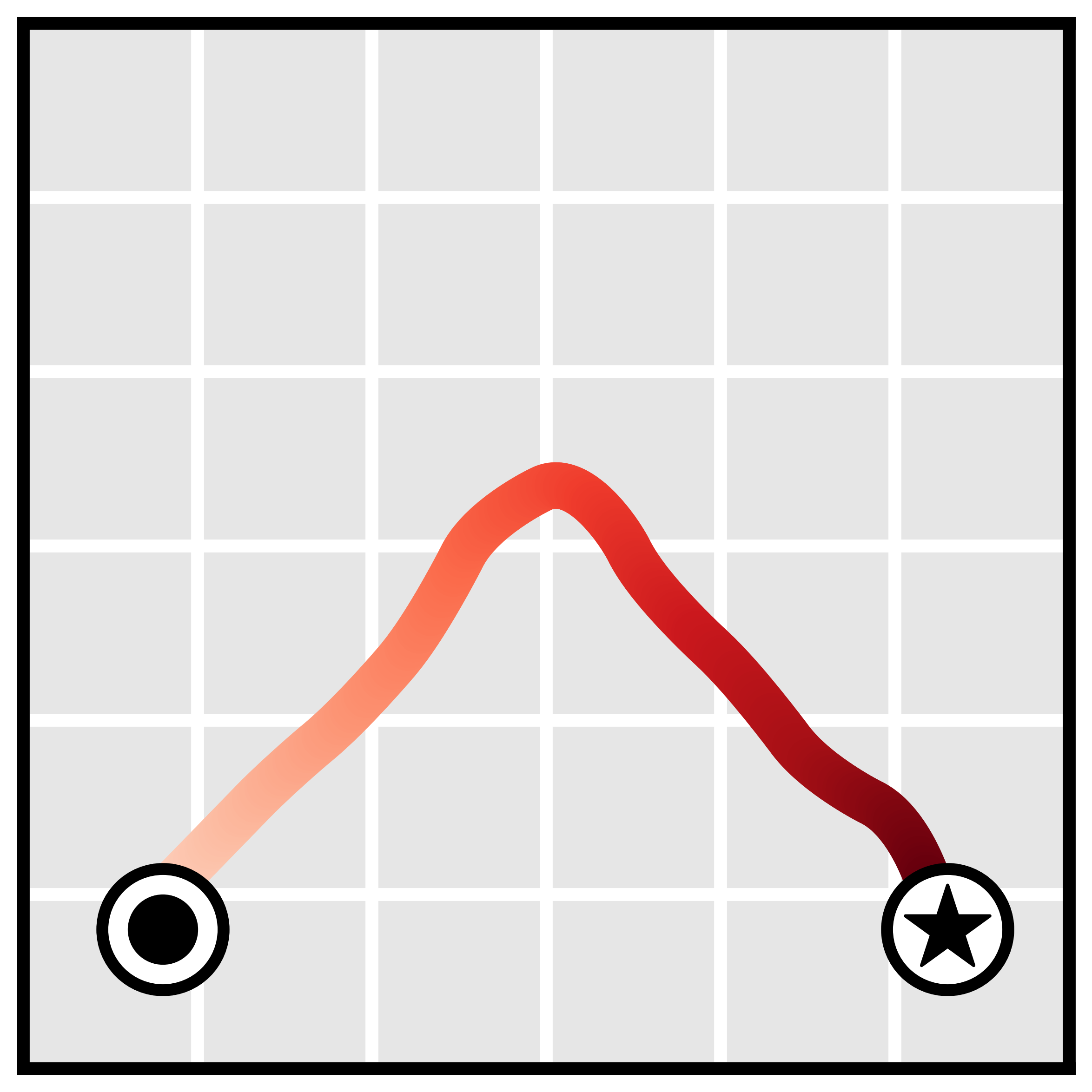} \\
    \vspace{-.5cm}
    \begin{center}
            \hspace{0.1cm} data \hspace{1.8cm} plan \\
    \end{center}
\end{minipage} \\
\vspace{.4cm}
\begin{minipage}[t]{0.55\textwidth}
    \raisebox{2cm}{\textbf{c}}~~~
    \raisebox{0.3cm}{\undersetimage{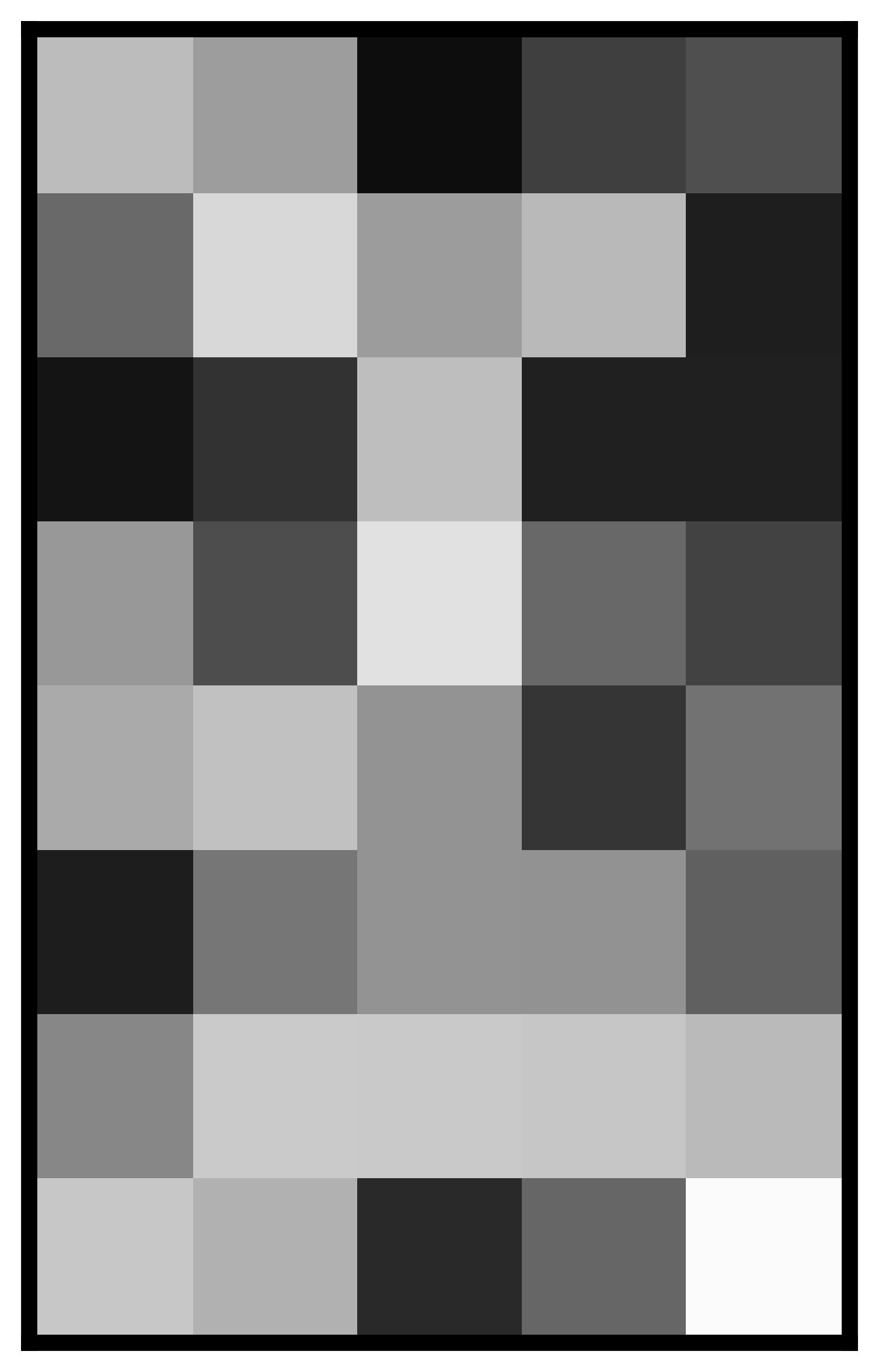}{$\btau{N}$}{-0.25}}
        \raisebox{1.2cm}{$\rightarrow$}
        \includegraphics[width=2.5cm]{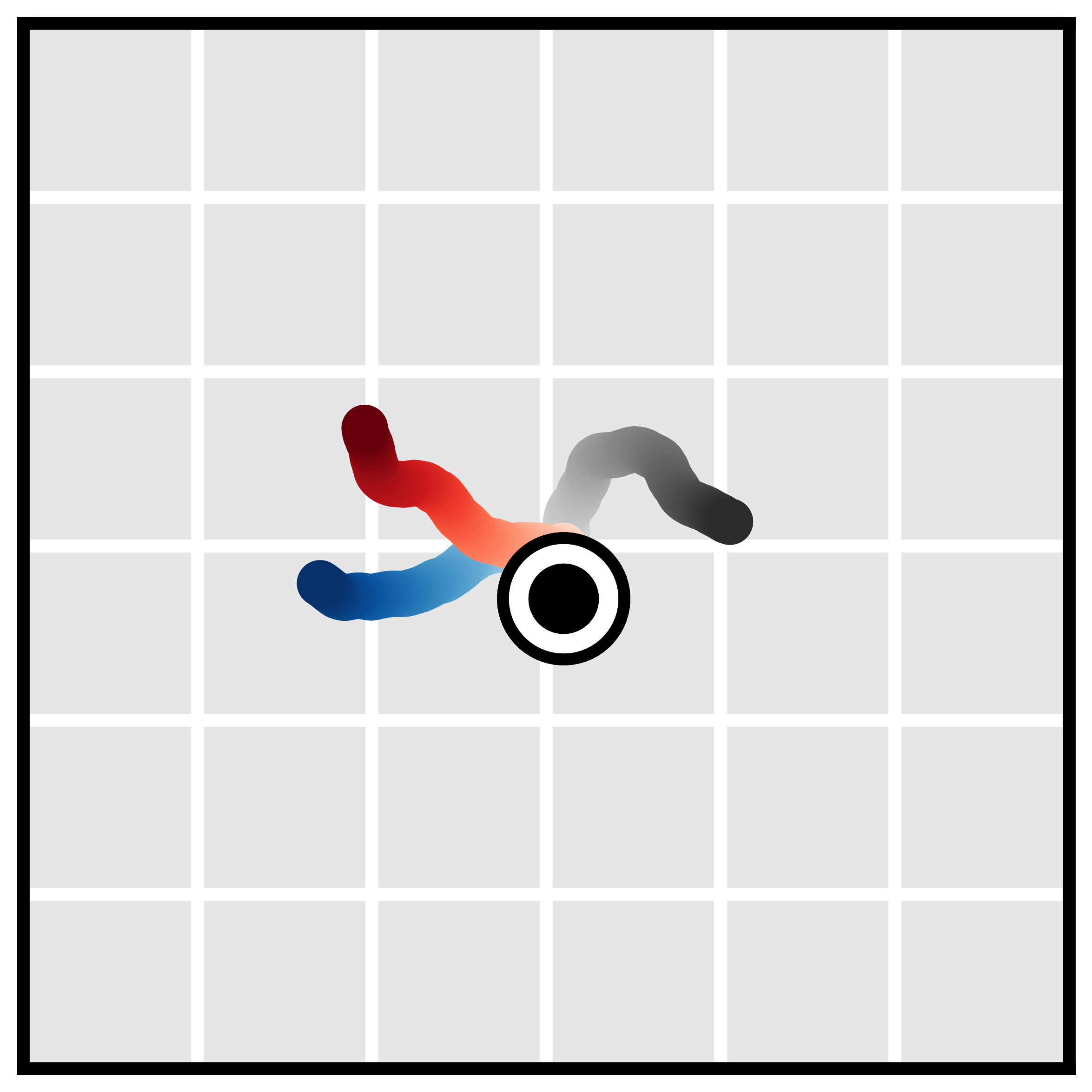}
        ~~
    \raisebox{-0.1cm}{\undersetimage{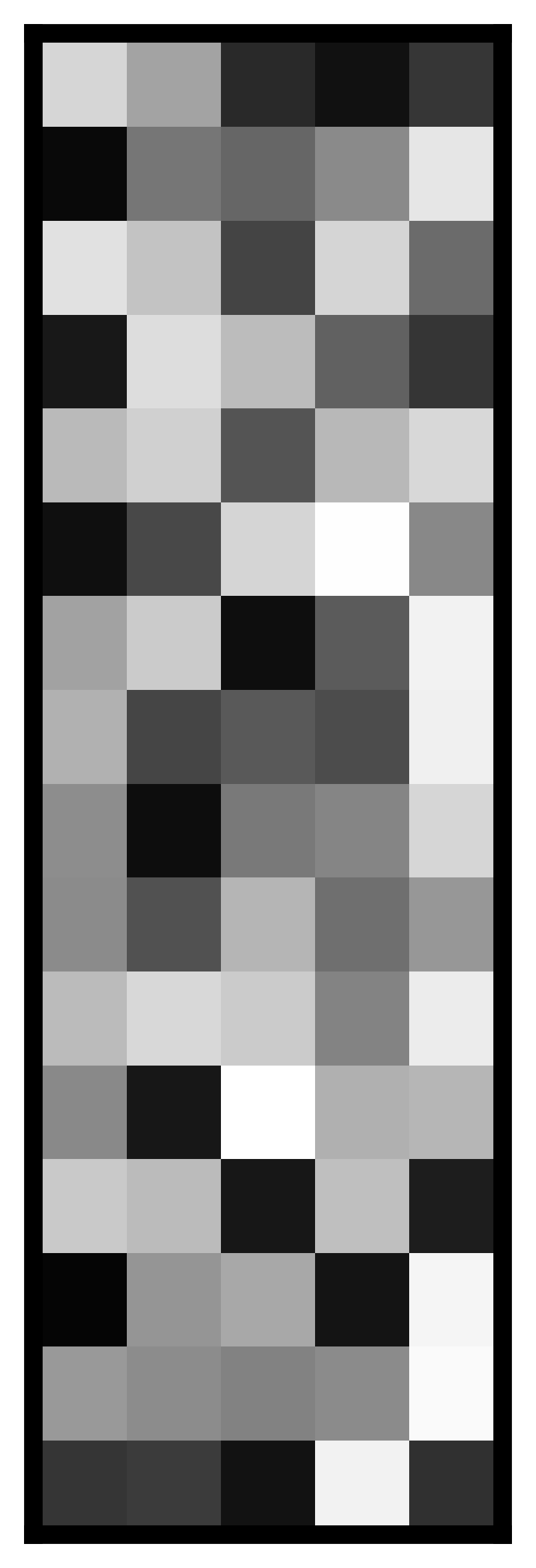}{$\btau{N}$}{-0.15}}
        \raisebox{1.2cm}{$\rightarrow$}
        \includegraphics[width=2.5cm]{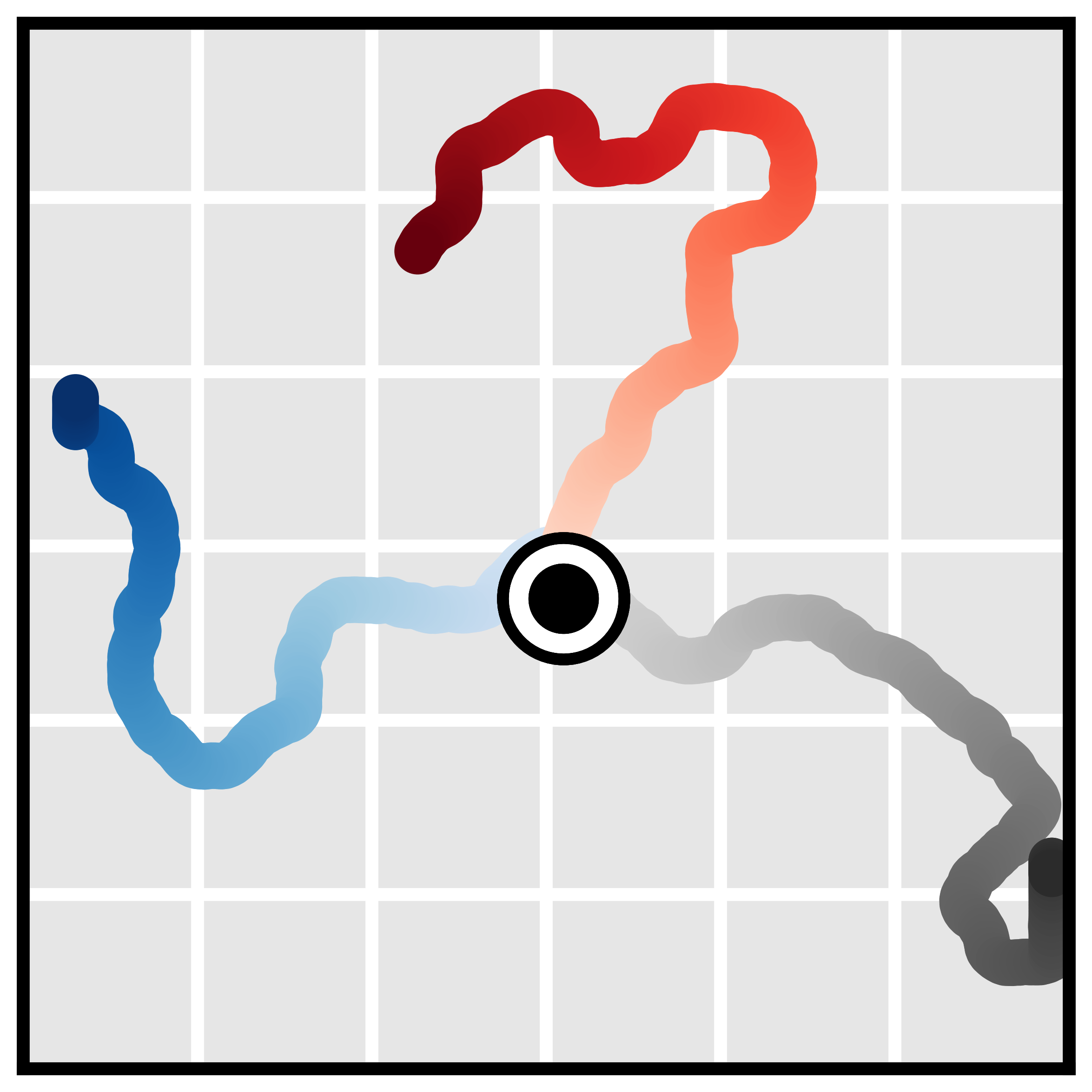} \\
\end{minipage}%
\begin{minipage}[t]{0.35\textwidth}
    \raisebox{2cm}{\textbf{d}}~~~
    \includegraphics[width=2.5cm]{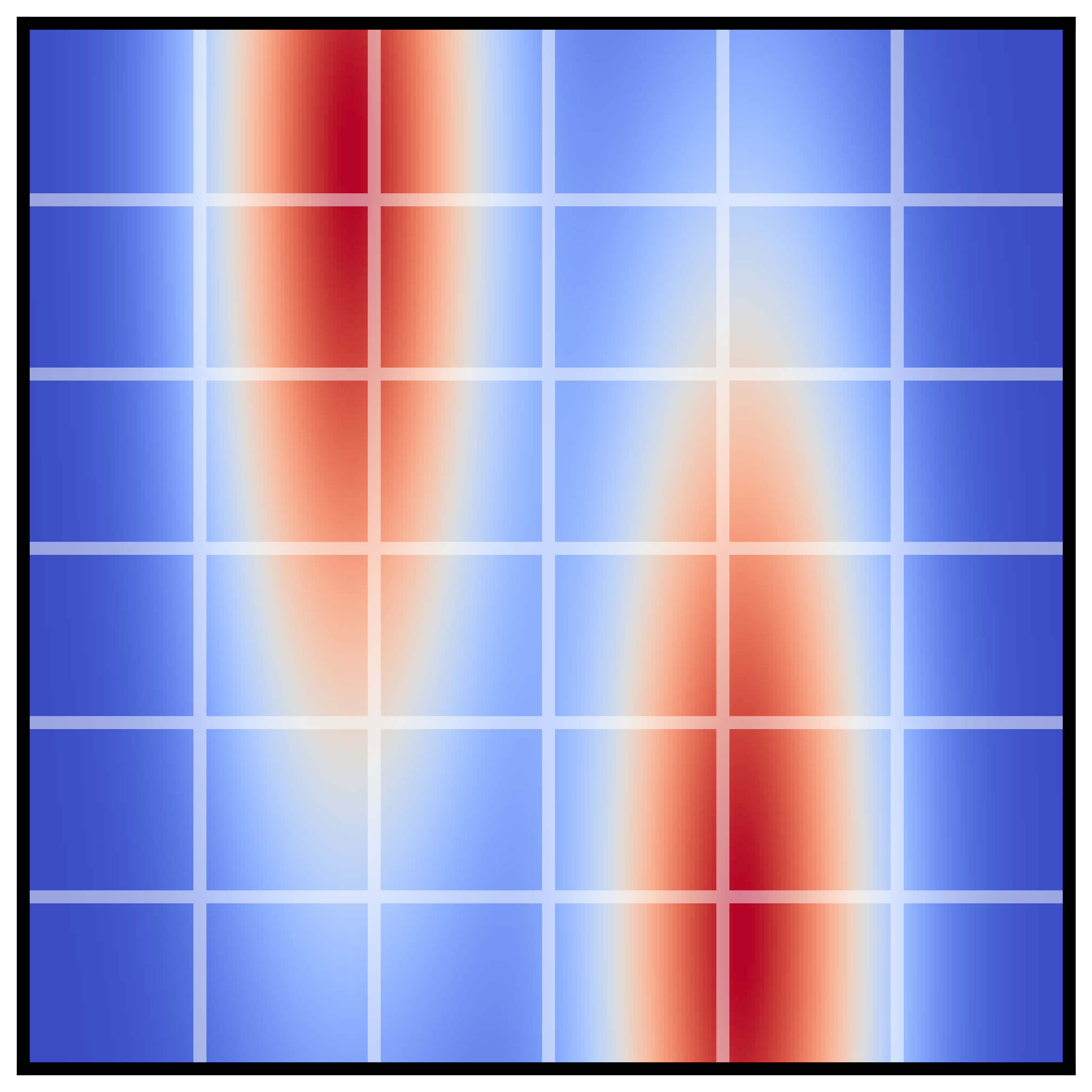}~
    \includegraphics[width=2.5cm]{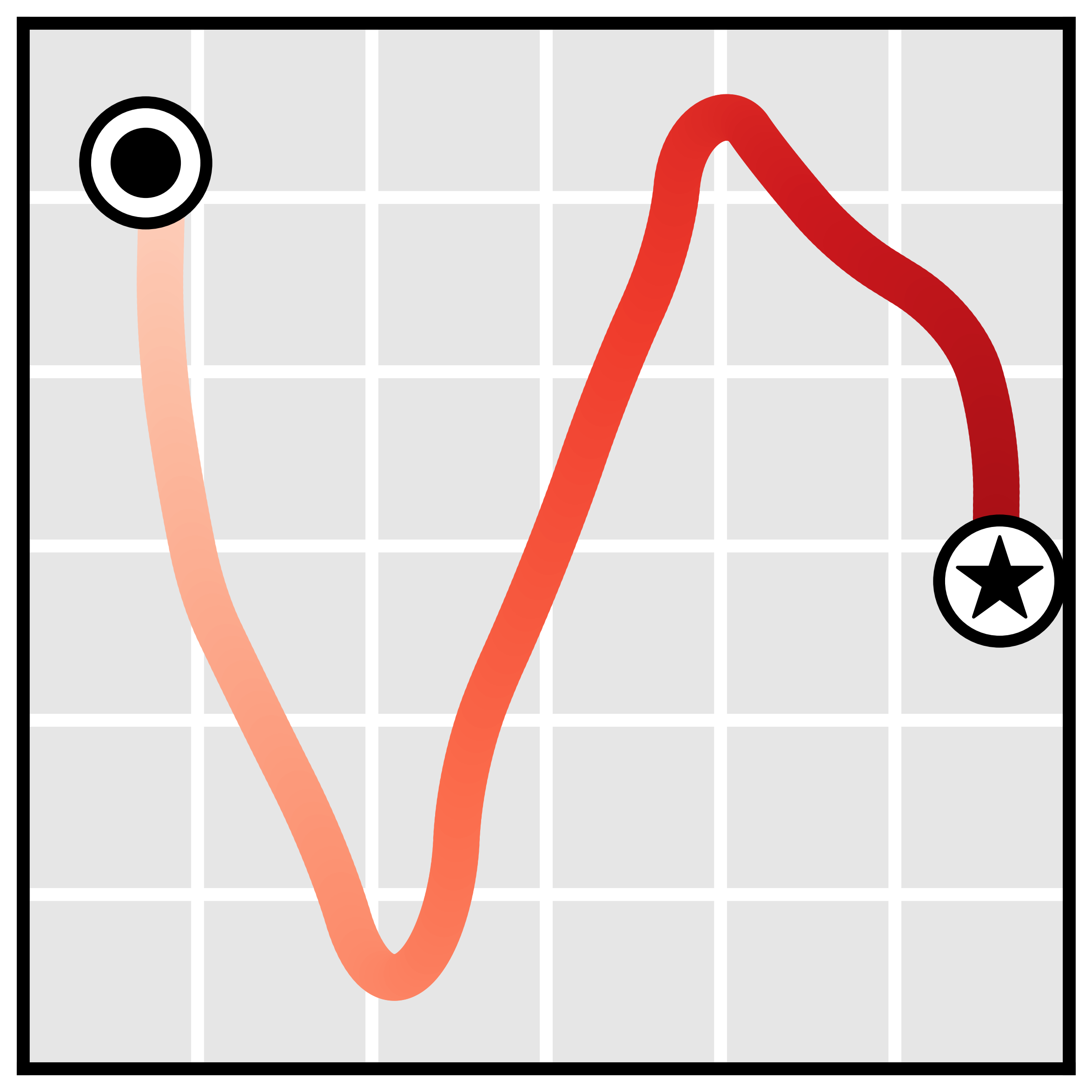} \\
    \vspace{-.5cm}
    \begin{flushleft}
            \hspace{0.43cm} reward
            \hspace{0.0cm}
            \raisebox{.05cm}{$-$}
            \raisebox{.0cm}{\includegraphics[height=1cm,angle=90]{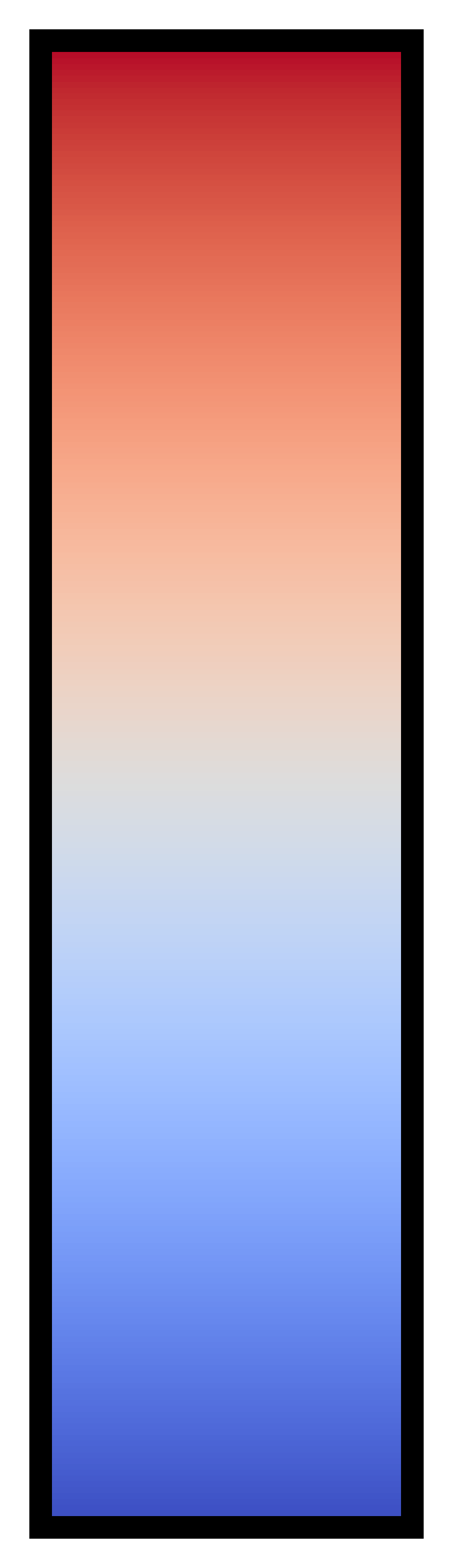}}
            \raisebox{.05cm}{$+$}
            \hspace{0.6cm} plan \\
    \end{flushleft}
\end{minipage}
\caption{
    \textbf{(Properties of diffusion planners)}
    \textbf{(a) Learned long-horizon planning:}
    \method{}'s learned planning procedure does not suffer from the myopic failure modes common to shooting algorithms and is able to plan over long horizons with sparse reward.
    \textbf{(b) Temporal compositionality:} Even though the model is not Markovian, it generates trajectories via iterated refinements to local consistency.
    As a result, it exhibits the types of generalization usually associated with Markovian models, with the ability to stitch together snippets of trajectories from the training data to generate novel plan.
    \textbf{(c) Variable-length plans:}
    Despite being a trajectory-level model, \method's planning horizon is not determined by its architecture.
    The horizon can be updated after training by changing the dimensionality of the input noise.
    \textbf{(d) Task compositionality:} \method{} can be composed with new reward functions to plan for tasks unseen during training.
    In all subfigures,
    \protect{\raisebox{-.05cm}{\includegraphics[height=.35cm]{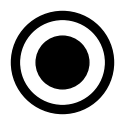}}}
    denotes a starting state and
    \protect{\raisebox{-.05cm}{\includegraphics[height=.35cm]{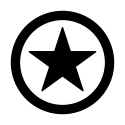}}}
    denotes a goal state.
}
\label{fig:properties}
\end{figure*}

\subsection{Reinforcement Learning as Guided Sampling}
\label{sec:guided}

In order to solve reinforcement learning problems with \method, we must introduce a notion of reward.
We appeal to the control-as-inference graphical model \citep{levine2018reinforcement} to do so.
Let $\mathcal{O}_{t}$ be a binary random variable denoting the optimality of timestep $t$ of a trajectory, with ${p(\mathcal{O}_t=1) = \exp(r(\st, \at))}$. We can sample from the set of optimal trajectories by setting ${h(\btau{}) = p(\mathcal{O}_{1:T} \mid \btau{})}$ in Equation~\ref{eq:perturbed}:
\[
\tilde{p}_\theta(\btau{}) = p(\btau{} \mid \mathcal{O}_{1:T}=1) \propto p(\btau{}) p(\mathcal{O}_{1:T}=1 \mid \btau{}).
\]

We have exchanged the reinforcement learning problem for one of \emph{conditional sampling}.
Thankfully, there has been much prior work on conditional sampling with diffusion models.
While it is intractable to sample from this distribution exactly, when $p(\mathcal{O}_{1:T} \mid \btau{i})$ is sufficiently smooth, the reverse diffusion process transitions can be approximated as Gaussian \citep{sohldickstein2015nonequilibrium}:
\begin{equation}
\label{eq:guided}
p_\theta(\btau{i-1} \mid \btau{i}, \opt_{1:T}) \approx \mathcal{N}(\btau{i-1}; \mu + \Sigma g, \Sigma)
\end{equation}

where $\mu, \Sigma$ are the parameters of the original reverse process transition $p_\theta(\btau{i-1} \mid \btau{i})$ and
\begin{align*}
g &= \nabla_{\btau{}} \log p(\opt_{1:T} \mid \btau{}) |_{\btau{} = \mu} \\
&= \sum_{t=0}^{T} \nabla_{\st,\at} r(\st, \at) |_{(\st,\at)=\mu_t}
= \nabla \mathcal{J}(\mu).
\end{align*}
\vspace{-.45cm}

This relation provides a straightforward translation between classifier-guided sampling, used to generate class-conditional images \citep{dhariwal2021diffusion}, and the reinforcement learning problem setting.
We first train a diffusion model $p_\theta(\btau{})$ on the states and actions of all available trajectory data.
We then train a separate model $\mathcal{J}_\phi$ to predict the cumulative rewards of trajectory samples $\btau{i}$.
The gradients of $\mathcal{J}_\phi$ are used to guide the trajectory sampling procedure by modifying the means $\mu$ of the reverse process according to Equation~\ref{eq:guided}.
The first action of a sampled trajectory ${\btau{} \sim p(\btau{} \mid \mathcal{O}_{1:T}=1)}$ may be executed in the environment, after which the planning procedure begins again in a standard receding-horizon control loop.
Pseudocode for the guided planning method is given in Algorithm~\ref{alg:rl}.

\subsection{Goal-Conditioned RL as Inpainting}
\label{sec:inpainting}

Some planning problems are more naturally posed as constraint satisfaction than reward maximization.
In these settings, the objective is to produce any feasible trajectory that satisfies a set of constraints, such as terminating at a goal location.
Appealing to the two-dimensional array representation of trajectories described by Equation~\ref{eq:trajectory_array}, this setting can be translated into an \emph{inpainting problem}, in which state and action constraints act analogously to observed pixels in an image \citep{sohldickstein2015nonequilibrium}.
All unobserved locations in the array must be filled in by the diffusion model in a manner consistent with the observed constraints.

The perturbation function required for this task is a Dirac delta for observed values and constant elsewhere.
Concretely, if $\mathbf{c}_t$ is state constraint at timestep $t$, then
\[
    h(\btau{}) = \delta_{\mathbf{c}_t}(\bs_0,\ba_0,\ldots,\bs_T,\ba_T) =
\begin{cases}
    +\infty      & \text{if } \mathbf{c}_t = \st\\
    ~~~~~0      & \text{otherwise}
\end{cases}
\]
The definition for action constraints is identical.
In practice, this may be implemented by sampling from the unperturbed reverse process ${\btau{i-1} \sim p_\theta(\btau{i-1} \mid \btau{i})}$ and replacing the sampled values with conditioning values $\mathbf{c}_t$ after all diffusion timesteps $i \in \{0, 1, \ldots, N\}$.

Even reward maximization problems require conditioning-by-inpainting because all sampled trajectories should begin at the current state.
This conditioning is described by line 10 in Algorithm~\ref{alg:rl}.

\section{Properties of Diffusion Planners}
\label{sec:properties}

We discuss a number of \method's important properties, focusing on those that are are either distinct from standard dynamics models or unusual for non-autoregressive trajectory prediction.


\textbf{Learned long-horizon planning.}~~
Single-step models are typically used as proxies for ground-truth environment dynamics $\vf$, and as such are not tied to any planning algorithm in particular.
In contrast, the planning routine in Algorithm~\ref{alg:rl} is closely tied to the specific affordances of diffusion models.
Because our planning method is nearly identical to sampling (with the only difference being guidance by a perturbation function $h(\btau{})$), \method's effectiveness as a long-horizon predictor directly translates to effective long-horizon planning.
We demonstrate the benefits of learned planning in a goal-reaching setting in \textbf{Figure~\ref{fig:properties}a}, showing that \method is able to generate feasible trajectories in the types of sparse reward settings where shooting-based approaches are known to struggle.
We explore a more quantitative version of this problem setting in Section~\ref{sec:maze2d}.

\vspace{.2em}
\textbf{Temporal compositionality.}~~
Single-step models are often motivated using the Markov property, allowing them to compose in-distribution transitions to generalize to out-of-distribution trajectories.
Because \method generates globally coherent trajectories by iteratively improving local consistency (Section~\ref{sec:model}), it can also stitch together familiar subsequences in novel ways.
In \textbf{Figure~\ref{fig:properties}b}, we train \method on trajectories that only travel in a straight line, and show that it can generalize to v-shaped trajectories by composing trajectories at their point of intersection.

\vspace{.2em}
\textbf{Variable-length plans.}~~
Because our model is fully convolutional in the horizon dimension of its prediction, its planning horizon is not specified by architectural choices.
Instead, it is determined by the size of the input noise $\btau{N} \sim \mathcal{N(\mathbf{0}, \mathbf{I})}$ that initializes the denoising process, allowing for variable-length plans (\textbf{Figure~\ref{fig:properties}c}).

\vspace{.2em}
\textbf{Task compositionality.}~~
While \method contains information about both environment dynamics and behaviors, it is independent of reward function.
Because the model acts as a prior over possible futures, planning can be guided by comparatively lightweight perturbation functions $h(\btau{})$ (or even combinations of multiple perturbations) corresponding to different rewards.
We demonstrate this by planning for a new reward function unseen during training of the diffusion model (\textbf{Figure~\ref{fig:properties}d}).


\begin{table}[t]
\centering
\small
\begin{tabular*}{\columnwidth}{@{\extracolsep{\fill}}lrrrrr}
\toprule
\multicolumn{1}{c}{\bf Environment} &
    \multicolumn{1}{c}{\bf MPPI} &
    \multicolumn{1}{c}{\bf CQL} &
    \multicolumn{1}{c}{\bf IQL} &
    \multicolumn{1}{c}{\bf \method} \\
\midrule
Maze2D~~~~U-Maze &
    $33.2$ &
    $5.7$ &
    $47.4$ &
    \highlight{\color{highlight}113.9} \scriptsize{\raisebox{1pt}{$\pm 3.1$}} \\ 
Maze2D~~~~Medium &
    $10.2$ &
    $5.0$ &
    $34.9$ &
    \highlight{\color{highlight}121.5} \scriptsize{\raisebox{1pt}{$\pm 2.7$}} \\ 
Maze2D~~~~Large &
    $5.1$ &
    $12.5$ &
    $58.6$ &
    \highlight{\color{highlight}123.0} \scriptsize{\raisebox{1pt}{$\pm 6.4$}} \\
\midrule
\multicolumn{1}{c}{\bf Single-task Average} & 16.2 & 7.7 & 47.0 & \highlight{\color{highlight}119.5} \hspace{.58cm} \\
\vspace{-.2cm} \\
\toprule
Multi2D~~~~U-Maze & 
    $41.2$ &
    - &
    $24.8$ &
    \highlight{\color{highlight}128.9} \scriptsize{\raisebox{1pt}{$\pm 1.8$}} \\ 
Multi2D~~~~Medium &
    $15.4$ &
    - &
    $12.1$ &
    \highlight{\color{highlight}127.2} \scriptsize{\raisebox{1pt}{$\pm 3.4$}} \\ 
Multi2D~~~~Large &
    $8.0$ &
    - &
    $13.9$ &
    \highlight{\color{highlight}132.1} \scriptsize{\raisebox{1pt}{$\pm 5.8$}} \\
\midrule
\multicolumn{1}{c}{\bf Multi-task Average} & 21.5 & - & 16.9 & \highlight{\color{highlight}129.4} \hspace{.58cm} \\ 
\bottomrule
\end{tabular*}
\caption{
    \textbf{(Long-horizon planning)}
    The performance of \method{} and prior model-free algorithms in the Maze2D environment, which tests long-horizon planning due to its sparse reward structure.
    The Multi2D setting refers to a multi-task variant with goal locations resampled at the beginning of every episode.
    \method{} substantially outperforms prior approaches in both settings.
    Appendix~\ref{app:baselines} details the sources for the scores of the baseline algorithms.
}
\label{table:maze2d}
\end{table}

\section{Experimental Evaluation}
\label{sec:experiments}

The focus of our experiments is to evaluate \method on the capabilities we would like from a data-driven planner.
In particular, we evaluate
\textbf{(1)} the ability to plan over long horizons without manual reward shaping,
\textbf{(2)} the ability to generalize to new configurations of goals unseen during training, and
\textbf{(3)} the ability to recover an effective controller from heterogeneous data of varying quality.
We conclude by studying practical runtime considerations of diffusion-based planning, including the most effective ways of speeding up the planning procedure while suffering minimally in terms of performance.

\subsection{Long Horizon Multi-Task Planning}
\label{sec:maze2d}

We evaluate long-horizon planning in the Maze2D environments \citep{fu2020d4rl}, which require traversing to a goal location where a reward of 1 is given.
No reward shaping is provided at any other location.
Because it can take hundreds of steps to reach the goal location, 
even the best model-free algorithms struggle to adequately perform credit assignment and reliably reach the goal (Table~\ref{table:maze2d}).

\begin{figure}[t!]
\begin{flushright}
    \raisebox{.35\height}{\rotatebox{90}{\textbf{U-Maze}}}\hspace*{\fill}
    \includegraphics[width=0.95\columnwidth]{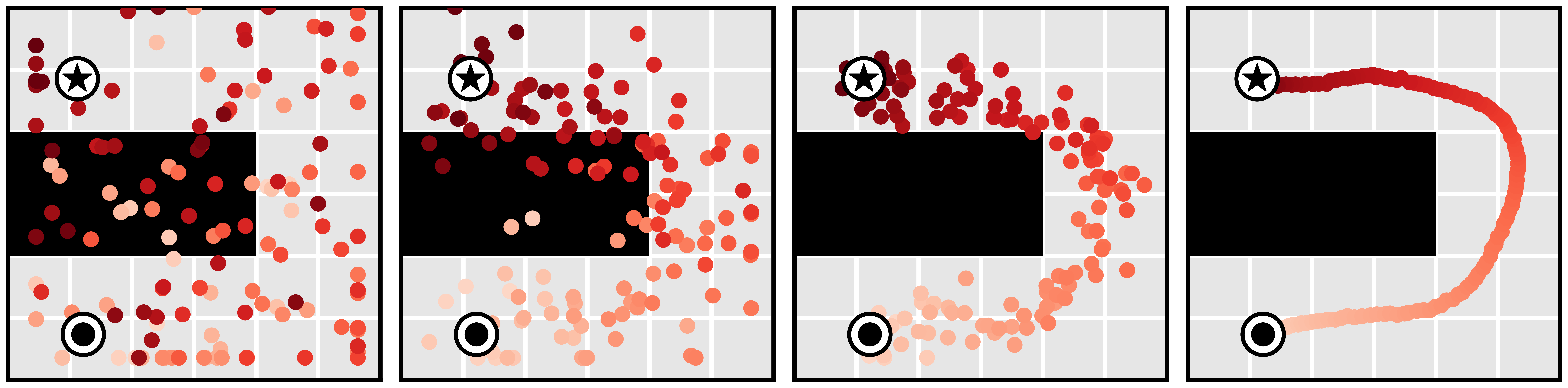} \\
    \raisebox{.3\height}{\rotatebox{90}{\textbf{Medium}}}\hspace*{\fill}
    \includegraphics[width=0.95\columnwidth]{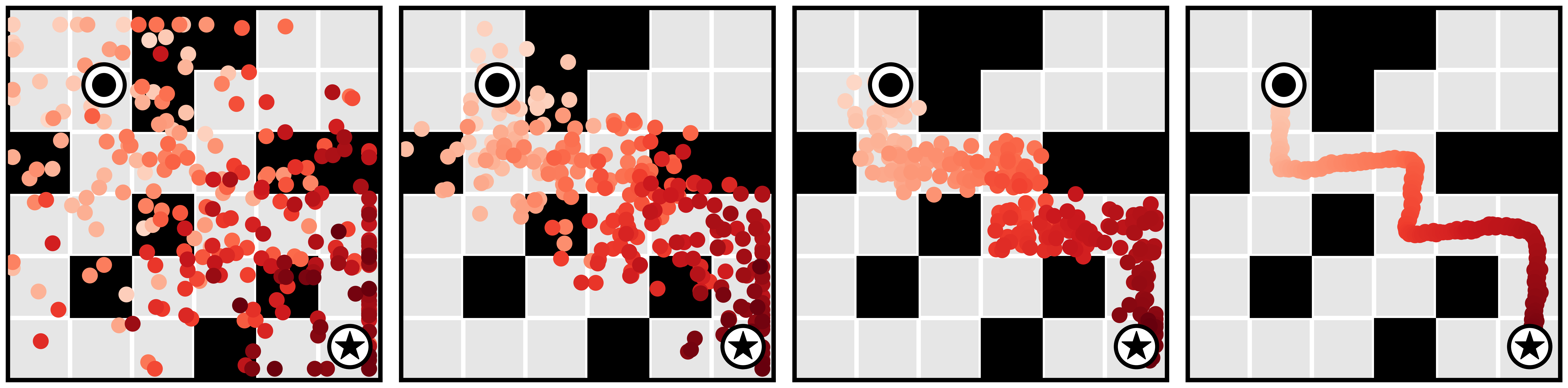} \\
    \raisebox{.6\height}{\rotatebox{90}{\textbf{Large}}}\hspace*{\fill}
    \includegraphics[width=0.95\columnwidth,trim={.03cm 0 .03cm 0},clip]{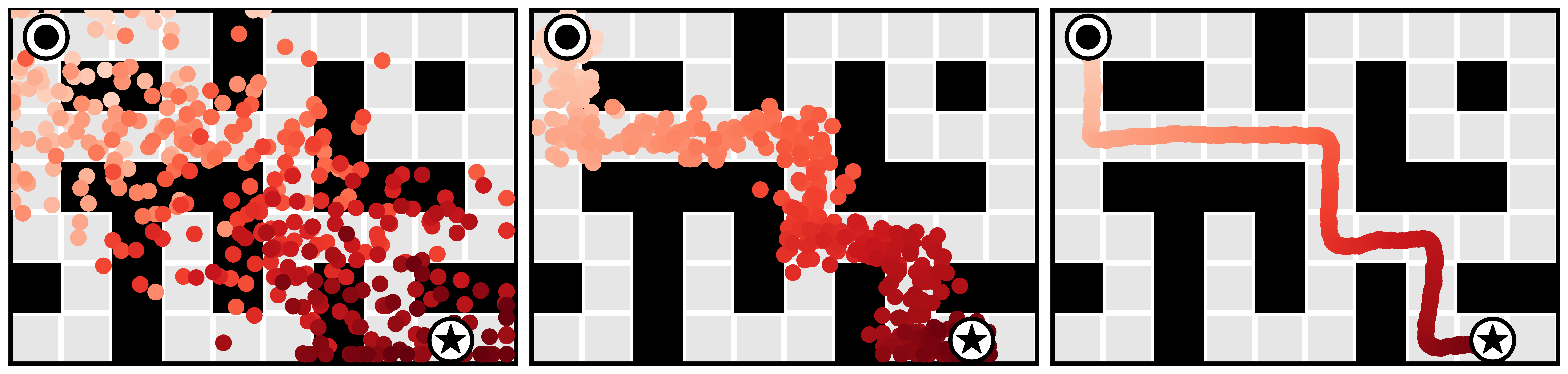} \\
\end{flushright}
\begin{centering}
    \raisebox{.3\height}{denoising}~~
    \tikz \draw [-{Computer Modern Rightarrow[length=2mm, width=3mm]}, line width=.3mm] (0,0) -- (3,0); \\
\end{centering}
\caption{
    \textbf{(Planning as inpainting)}
    Plans are generated in the Maze2D environment by sampling trajectories consistent with a specified start 
    \protect{\raisebox{-.05cm}{\includegraphics[height=.35cm]{images/maze2d/mark_start_crop.png}}}
    and goal
    \protect{\raisebox{-.05cm}{\includegraphics[height=.35cm]{images/maze2d/mark_goal_crop.png}}}
    condition.
    The remaining states are ``inpainted'' by the denoising process.
}
\label{fig:maze2d}
\end{figure}

We plan with \method using the inpainting strategy to condition on a start and goal location.
(The goal location is also available to the model-free methods; it is identifiable by being the only state in the dataset with non-zero reward.)
We then use the sampled trajectory as an open-loop plan.
\method achieves scores over 100 in all maze sizes, indicating that it outperforms a reference expert policy.
We visualize the reverse diffusion process generating \method's plans in Figure~\ref{fig:maze2d}.

While the training data in Maze2D is undirected -- consisting of a controller navigating to and from randomly selected locations -- the evaluation is single-task in that the goal is always the same.
In order to test multi-task flexibility, we modify the environment to randomize the goal location at the beginning of each episode.
This setting is denoted as Multi2D in Table~\ref{table:maze2d}.
\method{} is naturally a multi-task planner; we do not need to retrain the model from the single-task experiments and simply change the conditioning goal.
As a result, \method{} performs as well in the multi-task setting as in the single-task setting.
In contrast, there is a substantial performance drop of the best model-free algorithm in the single-task setting (IQL; \citealt{kostrikov2021implicit}) when adapted to the multi-task setting.
Details of our multi-task IQL with hindsight experience relabeling \citep{andrychowicz2017hindsight} are provided in Appendix~\ref{app:multitask}.
MPPI uses the ground-truth dynamics; its poor performance compared to the learned planning algorithm of Diffuser highlights the difficulty posed by long-horizon planning even when there are no prediction inaccuracies.

\definecolor{tblue}{HTML}{1F77B4}
\definecolor{tred}{HTML}{FF6961}
\definecolor{tgreen}{HTML}{429E9D}
\definecolor{thighlight}{HTML}{000000}
\newcolumntype{P}{>{\raggedleft\arraybackslash}X}
\begin{table*}[hb!]
\centering
\small
\begin{tabularx}{\textwidth}{llPPPPPPPPr}
\toprule
\multicolumn{1}{r}{\bf \color{black} Dataset} & \multicolumn{1}{r}{\bf \color{black} Environment} & \multicolumn{1}{r}{\bf \color{black} BC} & \multicolumn{1}{r}{\bf \color{black} CQL} & \multicolumn{1}{r}{\bf \color{black} IQL} & \multicolumn{1}{r}{\bf \color{black} DT} & \multicolumn{1}{r}{\bf \color{black} TT} & \multicolumn{1}{r}{\bf \color{black} MOPO} & \multicolumn{1}{r}{\bf \color{black} MOReL} & \multicolumn{1}{r}{\bf \color{black} MBOP} & \multicolumn{1}{r}{\bf \color{black} Diffuser} \\ 
\midrule
Medium-Expert & HalfCheetah & $55.2$ & $91.6$ & $86.7$ & $86.8$ & $95.0$ & $63.3$ & $53.3$ & $\textbf{\color{thighlight}105.9}$ & $88.9$ \scriptsize{\raisebox{1pt}{$\pm 0.3$}} \\ 
Medium-Expert & Hopper & $52.5$ & $\textbf{\color{thighlight}105.4}$ & $91.5$ & $\textbf{\color{thighlight}107.6}$ & $\textbf{\color{thighlight}110.0}$ & $23.7$ & $\textbf{\color{thighlight}108.7}$ & $55.1$ & $103.3$ \scriptsize{\raisebox{1pt}{$\pm 1.3$}} \\ 
Medium-Expert & Walker2d & $\textbf{\color{thighlight}107.5}$ & $\textbf{\color{thighlight}108.8}$ & $\textbf{\color{thighlight}109.6}$ & $\textbf{\color{thighlight}108.1}$ & $101.9$ & $44.6$ & $95.6$ & $70.2$ & $\textbf{\color{thighlight}106.9}$ \scriptsize{\raisebox{1pt}{$\pm 0.2$}} \\ 
\midrule
Medium & HalfCheetah & $42.6$ & $44.0$ & $\textbf{\color{thighlight}47.4}$ & $42.6$ & $\textbf{\color{thighlight}46.9}$ & $42.3$ & $42.1$ & $44.6$ & $42.8$ \scriptsize{\raisebox{1pt}{$\pm 0.3$}} \\ 
Medium & Hopper & $52.9$ & $58.5$ & $66.3$ & $67.6$ & $61.1$ & $28.0$ & $\textbf{\color{thighlight}95.4}$ & $48.8$ & $74.3$ \scriptsize{\raisebox{1pt}{$\pm 1.4$}} \\ 
Medium & Walker2d & $75.3$ & $72.5$ & $\textbf{\color{thighlight}78.3}$ & $74.0$ & $\textbf{\color{thighlight}79.0}$ & $17.8$ & $\textbf{\color{thighlight}77.8}$ & $41.0$ & $\textbf{\color{thighlight}79.6}$ \scriptsize{\raisebox{1pt}{$\pm 0.55$}} \\ 
\midrule
Medium-Replay & HalfCheetah & $36.6$ & $45.5$ & $44.2$ & $36.6$ & $41.9$ & $\textbf{\color{thighlight}53.1}$ & $40.2$ & $42.3$ & $37.7$ \scriptsize{\raisebox{1pt}{$\pm 0.5$}} \\ 
Medium-Replay & Hopper & $18.1$ & $\textbf{\color{thighlight}95.0}$ & $\textbf{\color{thighlight}94.7}$ & $82.7$ & $\textbf{\color{thighlight}91.5}$ & $67.5$ & $\textbf{\color{thighlight}93.6}$ & $12.4$ & $\textbf{\color{thighlight}93.6}$ \scriptsize{\raisebox{1pt}{$\pm 0.4$}} \\ 
Medium-Replay & Walker2d & $26.0$ & $77.2$ & $73.9$ & $66.6$ & $\textbf{\color{thighlight}82.6}$ & $39.0$ & $49.8$ & $9.7$ & $70.6$ \scriptsize{\raisebox{1pt}{$\pm 1.6$}} \\ 
\midrule
\multicolumn{2}{c}{\bf Average} & 51.9 & \textbf{\color{thighlight}77.6} & \textbf{\color{thighlight}77.0} & 74.7 & \textbf{\color{thighlight}78.9} & 42.1 & 72.9 & 47.8 & \textbf{\color{thighlight}77.5} \hspace{.6cm} \\ 
\bottomrule
\end{tabularx}
\vspace{-.0cm}
\caption{
    \textbf{(Offline reinforcement learning)}
    The performance of \method{} and a variety of prior algorithms on the D4RL locomotion benchmark \citep{fu2020d4rl}.
    Results for \method{} correspond to the mean and standard error over 150 planning seeds. We detail the sources for the performance of prior methods in Appendix~\ref{app:d4rl_sources}. Following \citet{kostrikov2021implicit}, we emphasize in bold scores within 5 percent of the maximum per task ($\ge 0.95 \cdot \text{max}$).
}
\label{table:locomotion}
\end{table*}

\vspace{-.2cm}
\subsection{Test-time Flexibility}
\label{sec:blocks}

In order to evaluate the ability to generalize to new test-time goals, we construct a suite of block stacking tasks with three settings: \textbf{(1)} {\it Unconditional Stacking}, for which the task is to build a block tower as tall as possible; \textbf{(2)} {\it Conditional Stacking}, for which the task is to construct a block tower with a specified order of blocks, and \textbf{(3)} {\it Rearrangement}, for which the task is to match a set of reference blocks' locations in a novel arrangement.
We train all methods on 10000 trajectories from demonstrations generated by PDDLStream \citep{garrett2020pddlstream}; rewards are equal to one upon successful stack placements and zero otherwise.
These block stacking are challenging diagnostics of test-time flexibility;
in the course of executing a partial stack for a randomized goal, a controller will venture into novel states not included in the training configuration.

We use one trained \method for all block-stacking tasks, only modifying the perturbation function $h(\btau{})$ between settings.
In the {\it Unconditional Stacking} task, we directly sample from the unperturbed denoising process $p_\theta(\btau{})$ to emulate the PDDLStream controller.
In the {\it Conditional Stacking} and {\it Rearrangement} tasks, we compose two perturbation functions $h(\btau{})$ to bias the sampled trajectories: the first maximizes the likelihood of the trajectory's final state matching the goal configuration, and the second enforces a contact constraint between the end effector and a cube during stacking motions.
(See Appendix~\ref{app:stacking_costs} for details.)

\begin{figure}[t]
    \centering
    \small
    \begin{tabular}{@{\extracolsep{\fill}}lrrr}
    \toprule
    \multicolumn{1}{c}{\bf Environment} & \multicolumn{1}{c}{\bf ~~~~~BCQ~} &
        \multicolumn{1}{c}{\bf ~~~~~CQL~} &
        \multicolumn{1}{c}{\bf ~~\method} \\
    \midrule
    Unconditional Stacking & $0.0$ & $24.4$ & ~\highlight{\color{highlight}58.7} \scriptsize{\raisebox{1pt}{$\pm 2.5$}}
         \\ 
    Conditional Stacking & 0.0  & 0.0 & \highlight{\color{highlight}45.6} \scriptsize{\raisebox{1pt}{$\pm 3.1$}}
         \\ 
    Rearrangement & 0.0  & 0.0 & \highlight{\color{highlight}58.9} \scriptsize{\raisebox{1pt}{$\pm 3.4$}} \\ 
    \midrule
    \multicolumn{1}{c}{\bf Average} & 0.0  & 8.1 & \highlight{\color{highlight}54.4~~} \hspace{.44cm} \\ 
    \bottomrule
    \end{tabular}
    \captionof{table}{
    \textbf{(Test-time flexibility)}
    Performance of BCQ, CQL, and \method on block stacking tasks.
    A score of 100 corresponds to a perfectly executed stack; 0 is that of a random policy.
    }
    \label{table:blocks}
    \vspace{-.6cm}
\end{figure}

\begin{figure}[t]
    \vspace{-.25cm}
    \centering
    \includegraphics[width=0.45\columnwidth]{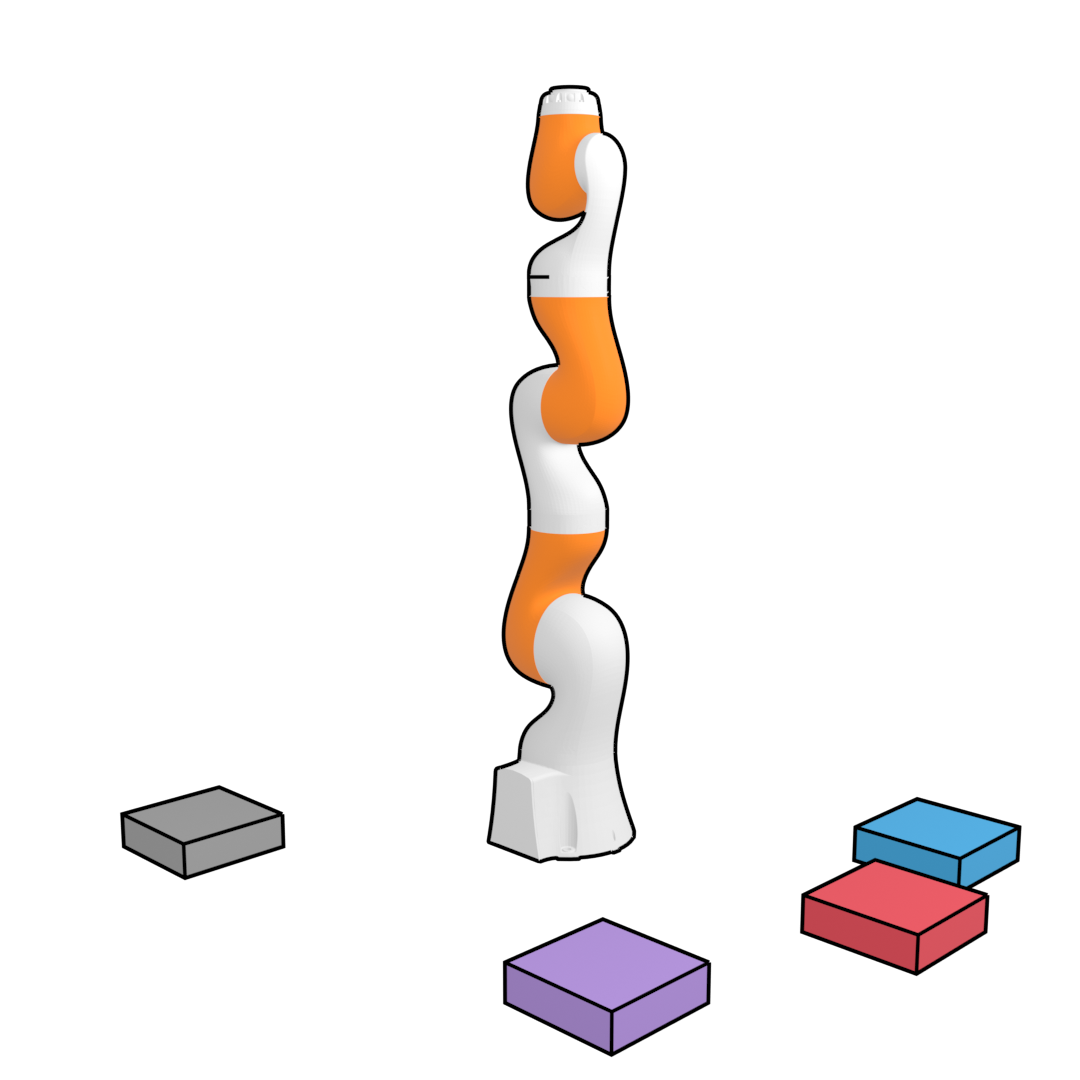}
    \raisebox{1.7cm}{\hspace{-.1cm}$\rightarrow$}\hspace{.3cm}
    \includegraphics[width=0.45\columnwidth]{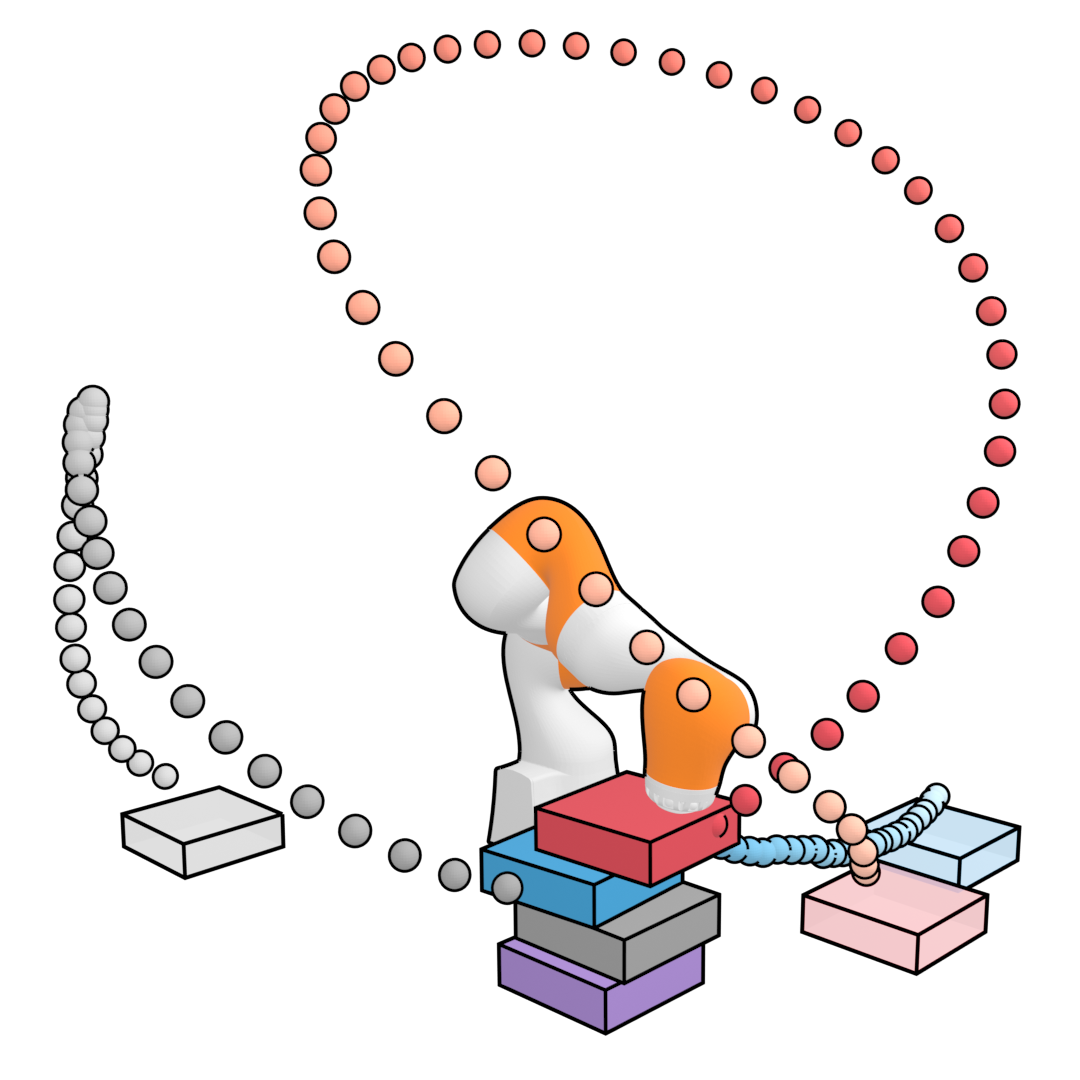}
    \caption{
        \textbf{(Block stacking)}
        A block stacking sequence executed by \method. This task is best illustrated by videos viewable at
        \href{https://diffusion-planning.github.io/}
        {\texttt{diffusion-planning.github.io}}.
    }
    \label{fig:blocks}
    \vspace{-.3cm}
\end{figure}

We compare with two prior model-free offline reinforcement learning algorithms: BCQ \citep{fujimoto2019off} and CQL \citep{kumar2020conservative}, training standard variants for {\it Unconditional Stacking} and goal-conditioned variants for {\it Conditional Stacking} and {\it Rearrangement}.
(Baseline details are provided in Appendix~\ref{app:baselines}.)
Quantitative results are given in Table~\ref{table:blocks}, in which a score of 100 corresponds to a perfect execution of the task.
\method substantially outperforms both prior methods, with the conditional settings requiring flexible behavior generation proving especially difficult for the model-free algorithms.
A visual depiction of an execution by \method is provided in Figure~\ref{fig:blocks}.

\subsection{Offline Reinforcement Learning}
\label{sec:locomotion}
\vspace{-.1cm}

Finally, we evaluate the capacity to recover an effective single-task controller from heterogeneous data of varying quality using the D4RL offline locomotion suite \citep{fu2020d4rl}.
We guide the trajectories generated by \method toward high-reward regions using the sampling procedure described in Section~\ref{sec:guided} and condition the trajectories on the current state using the inpainting procedure described in Section~\ref{sec:inpainting}.
The reward predictor $\mathcal{J}_\phi$ is trained on the same trajectories as the diffusion model.

We compare to a variety of prior algorithms spanning other approaches to data-driven control, including the model-free reinforcement learning algorithms CQL \citep{kumar2020conservative} and IQL \citep{kostrikov2021implicit}; return-conditioning approaches like Decision Transformer (DT; \citealt{chen2021decision}); and model-based reinforcement learning approaches including Trajectory Transformer (TT; \citealt{janner2021sequence}), MOPO \citep{yu2020mopo}, MOReL \citep{kidambi2020morel}, and MBOP \citep{argenson2020model}.
In the single-task setting, \method performance comparably to prior algorithms: better than the model-based MOReL and MBOP and return-conditioning DT, but worse than the best offline techniques designed specifically for single-task performance.
We also investigated a variant using Diffuser as a dynamics model in conventional trajectory optimizers such as MPPI \citep{grady2015mppi}, but found that this combination performed no better than random, suggesting that the effectiveness of Diffuser stems from coupled modeling and planning, and not from improved open-loop predictive accuracy.

\begin{figure}
\centering
\begin{minipage}{.05\columnwidth}
\begin{flushleft}
    \raisebox{.2\height}{\rotatebox{90}{denoising}}
    \tikz \draw [-{Computer Modern Rightarrow[length=2mm, width=3mm]}, line width=.3mm] (0,2) -- (0,0); \\
\end{flushleft}
\end{minipage}~~
\begin{minipage}{.9\columnwidth}
\centering
    \vspace{.15cm}
    \includegraphics[width=\linewidth]{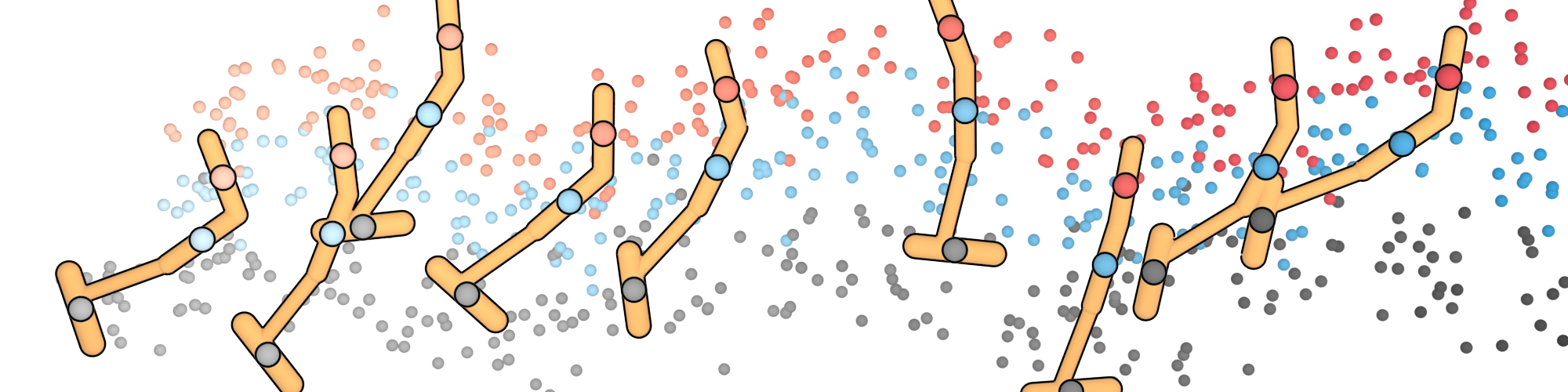} \\
    \includegraphics[width=\linewidth]{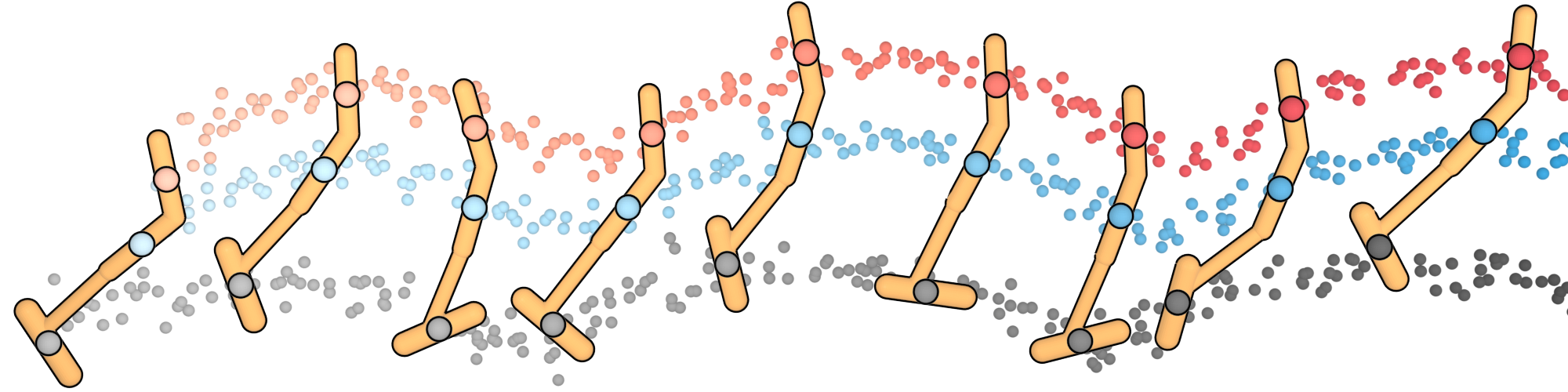} \\
    \includegraphics[width=\linewidth]{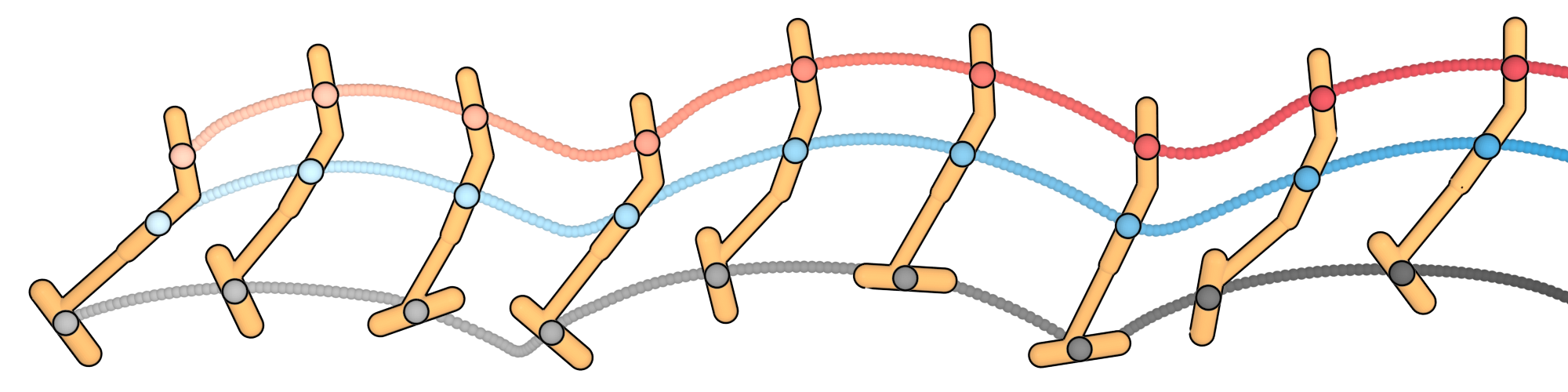} \\
\end{minipage} \\
\vspace{.3cm}
\hspace{1cm}
\raisebox{.3\height}{planning horizon}~~
\tikz \draw [-{Computer Modern Rightarrow[length=2mm, width=3mm]}, line width=.3mm] (0,0) -- (3,0); \\
\caption{
    \textbf{(Guided sampling)}
    \method generates all timesteps of a plan concurrently, instead of autoregressively, through the denoising process.
}
\vspace{-15pt}
\label{fig:locomotion}
\end{figure}

\subsection{Warm-Starting Diffusion for Faster Planning}

A limitation of \method{} is that individual plans are slow to generate (due to iterative generation). Na\"ively,  as we execute plans open loop, a new plan must be regenerated at each step of execution. To improve execution speed of \method{}, we may further reuse previously generated plans to warm-start generations of subsequent plans. 

\begin{figure}
    \centering
    \includegraphics[width=\columnwidth]{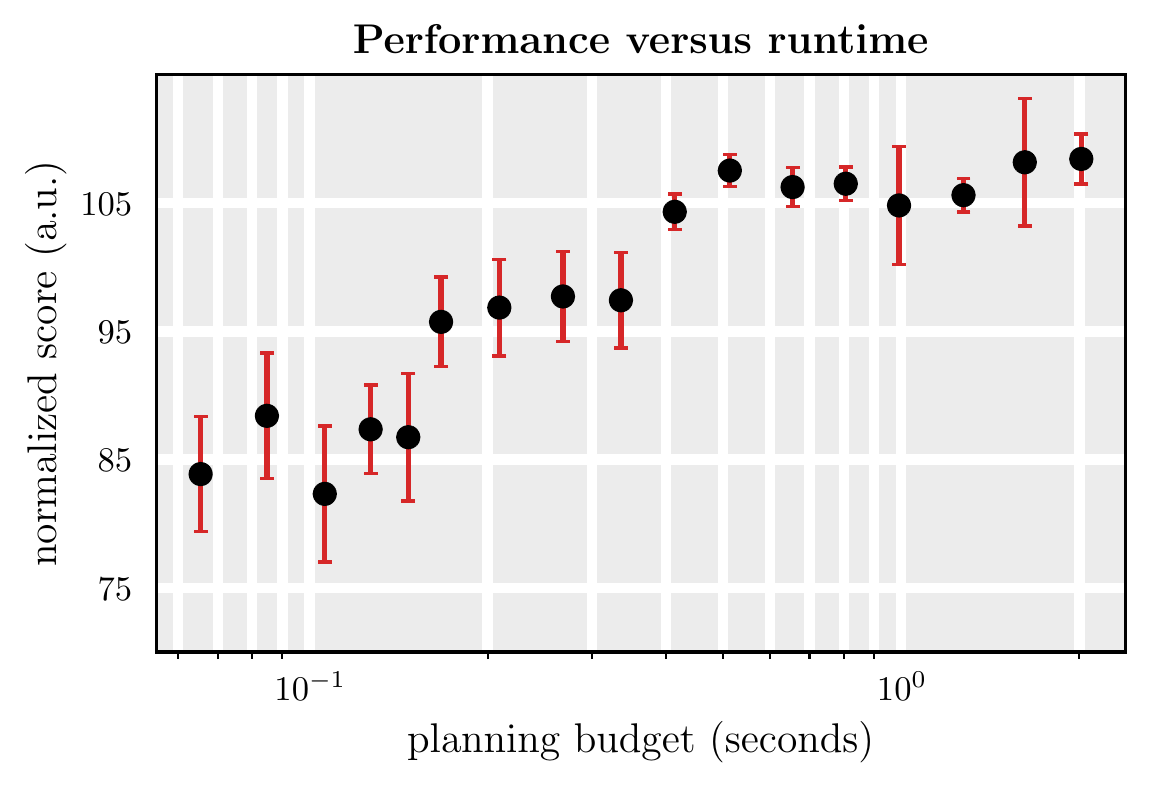}
    \vspace{-.25cm}
    \caption{
    \textbf{(Fast Planning)}
    Performance of \method on Walker2d Medium-Expert when varying the number of diffusion steps to warm-start planning.
    Performance suffers only minimally even when using one-tenth the number of diffusion steps, as long as plans are initialized from the previous timestep's plan.}
    \label{fig:speed}
    \vspace{-5pt}
\end{figure}

To warm-start planning, we may run a limited number of forward diffusion steps from a previously generated plan and then run a corresponding number of denoising steps from this partially noised trajectory to regenerate an updated plan. In \fig{fig:speed}, we illustrate the trade-off between performance and runtime budget as we vary the underlying number of denoising steps used to regenerate each a new plan from 2 to 100. We find that we may reduce the planning budget of our approach markedly with only modest drop in performance.

\vspace{-.15cm}
\section{Related Work}

Advances in deep generative modeling have recently made inroads into model-based reinforcement learning, with multiple lines of work exploring dynamics models parameterized as
convolutional U-networks \cite{kaiser2019mbatari},
stochastic recurrent networks \cite{ke2018modeling,hafner2018planet,ha2018worldmodels},
vector-quantized autoencoders \cite{hafner2021mastering,ozair2021vector},
neural ODEs \cite{du2020ode},
normalizing flows \cite{rhinehart2020imitative,janner2020gamma},
generative adversarial networks \cite{eysenbach2021mismatched},
energy-based models (\emph{EBMs}; \citealt{du2019model}),
graph neural networks  \cite{sanchez-gonzalez2018graph},
neural radiance fields \cite{li2021scene},
and Transformers \citep{janner2021sequence,chen2021transdreamer}.
Further, \citealt{lambert2020learning} have studied non-autoregressive trajectory-level dynamics models for long-horizon prediction.
These investigations generally assume an abstraction barrier between the model and planner.
Specifically, the role of learning is relegated to approximating environment dynamics; once learning is complete the model may be inserted into any of a variety of planning \cite{botev2013cem,grady2015mppi} or policy optimization \cite{sutton1990dyna,wang2019benchmarking} algorithms because the form of the planner does not depend strongly on the form of the model.
Our goal is to break this abstraction barrier by designing a model and planning algorithm that are trained alongside one another,
resulting in a non-autoregressive trajectory-level model for which sampling and planning are nearly identical.

A number of parallel lines of work have studied how to break the abstraction barrier between model learning and planning in different ways.
Approaches include training an autoregressive latent-space model for reward prediction \citep{tamar2016vin, oh2017vpn,schrittwieser2019muzero}; weighing model training objectives by state values \citep{farahmand2017valueaware}; and applying collocation techniques to learned single-step energies \citep{du2019model, rybkin2021model}.
In contrast, our method plans by modeling and generating all timesteps of a trajectory concurrently, instead of autoregressively, and conditioning the sampled trajectories with auxiliary guidance functions.


Diffusion models have emerged as a promising class of generative model that formulates the data-generating process as an iterative denoising procedure \citep{sohldickstein2015nonequilibrium, ho2020denoising}.
The denoising procedure can be seen as parameterizing the gradients of the data distribution \citep{song2019generative}, connecting diffusion models to score matching \cite{hyvarinen2005score} and EBMs \citep{lecun06atutorial, du2019implicit, nijkamp2019learning,grathwohl2020stein}.
Iterative, gradient-based sampling lends itself towards flexible conditioning \citep{dhariwal2021diffusion} and compositionality \citep{du2020compositional}, which we use to recover effective behaviors from heterogeneous datasets and plan for reward functions unseen during training.
While diffusion models have been developed for the generation of images \cite{song2020denoising}, waveforms \cite{chen2020wavegrad}, 3D shapes \cite{zhou2021shape}, and text \cite{austin2021structured}, to the best of our knowledge they have not previously been used in the context of reinforcement learning or decision-making.
\vspace{-.15cm}
\section{Conclusion}
\label{sec:discussion}
We have presented \method, a denoising diffusion model for trajectory data.
Planning with \method is almost identical to sampling from it, differing only in the addition of auxiliary perturbation functions that serve to guide samples.
The learned diffusion-based planning procedure has a number of useful properties, including graceful handling of sparse rewards, the ability to plan for new rewards without retraining, and a temporal compositionality that allows it to produce out-of-distribution trajectories by stitching together in-distribution subsequences.
Our results point to a new class of diffusion-based planning procedures for deep model-based reinforcement learning.

\subsection*{Code References}
We used the following open-source libraries for this work: NumPy \citep{harris2020numpy}, PyTorch \citep{paszke2019pytorch}, and Diffusion Models in PyTorch \citep{wang2020ddpm}.

\subsection*{Acknowledgements}
We thank Ajay Jain for feedback on an early draft and Leslie Kaelbling, Tom\'as Lozano-P\'erez, Jascha Sohl-Dickstein, Ben Eysenbach, Amy Zhang, Colin Li, and Toru Lin for helpful discussions.
This work was partially supported by computational resource donations from Microsoft.
M.J. is supported by fellowships from the National Science Foundation and the Open Philanthropy Project.
Y.D. is supported by a fellowship from the National Science Foundation.

\bibliography{ms}
\bibliographystyle{setup/icml2022}

\clearpage

\titleformat{\section}
  {\normalfont\large\bfseries}{Appendix \thesection}{1em}{}

\renewcommand\thefigure{A\arabic{figure}} 
\setcounter{figure}{0} 

\begin{appendices}

\section{Baseline details and sources}
\label{app:baselines}
\label{app:multitask}

In this section, we provide details about baselines we ran ourselves.
For scores of baselines previously evaluated on standardized tasks, we provide the source of the listed score.

\subsection{Maze2D experiments}
        
\paragraph{Single-task.}

The performance of CQL and IQL on the standard Maze2D environments is reported in the D4RL whitepaper \citep{fu2020d4rl} in Table~2.

We ran IQL using the offical implementation from the authors: 
\begin{center}
{
    \small
    \href{https://github.com/ikostrikov/implicit_q_learning}
    {\texttt{github.com/ikostrikov/implicit\_q\_learning}}.
}
\end{center}

We tuned over two hyperparameters:
\begin{enumerate}
    \item temperature $\in [3, 10]$
    \item expectile $\in [0.65, 0.95]$
\end{enumerate}

\paragraph{Multi-task.}

We only evaluated IQL on the Multi2D environments because it is the strongest baseline in the single-task Maze2D environments by a sizeable margin.
To adapt IQL to the multi-task setting, we modified the $Q$-functions, value function, and policy to be goal-conditioned.
To select goals during training, we employed a strategy based on hindsight experience replay, in which we sampled a goal from among those states encountered in the future of a trajectory.
For a training backup $(\st, \at, \stp)$, we sampled goals according to a geometric distribution over the future
\[
\Delta \sim \text{Geom}(1-\gamma) ~~~~~~~~ \mathbf{g} = \bs_{t + \Delta},
\]
recalculated rewards based on the sampled goal, and conditioned all relevant models on the goal during updating.
During testing, we conditioned the policy on the ground-truth goal.

We tuned over the same IQL parameters as in the single-task setting.

\subsection{Block stacking experiments}

\paragraph{Single-task.}

We ran CQL using the following implementation
\begin{center}
{
    \small
    \href{https://github.com/young-geng/cql}
    {\texttt{https://github.com/young-geng/cql}}.
}
\end{center}
and used default hyperparameters in the code.
We ran BCQ using the author's original implementation
\begin{center}
{
    \small
    \href{https://github.com/sfujim/BCQ}
    {\texttt{https://github.com/sfujim/BCQ}}.
}
\end{center}

For BCQ, we tuned over two hyperparameters:
\begin{enumerate}
    \item discount factor $\in [0.9, 0.999]$
    \item tau $\in [0.001, 0.01]$
\end{enumerate}

\paragraph{Multi-task.}

To evaluate BCQ and  CQL in the multi-task setting, we modified the $Q$-functions, value function and policy to be goal-conditioned.
We trained using goal relabeling as in the Multi2D environments.
We tuned over the same hyperparameters described in the single-task block stacking experiments.




\subsection{Offline Locomotion}
\label{app:d4rl_sources}

The scores for BC, CQL, IQL, and AWAC are from Table~1 in \citet{kostrikov2021implicit}.
The scores for DT are from Table~2 in \citet{chen2021decision}.
The scores for TT are from Table~1 in \citet{janner2021sequence}.
The scores for MOReL are from Table~2 in \citet{kidambi2020morel}.
The scores for MBOP are from Table~1 in \citet{argenson2020model}.

\begin{figure}[t]
\centering
\includegraphics[width=\columnwidth]{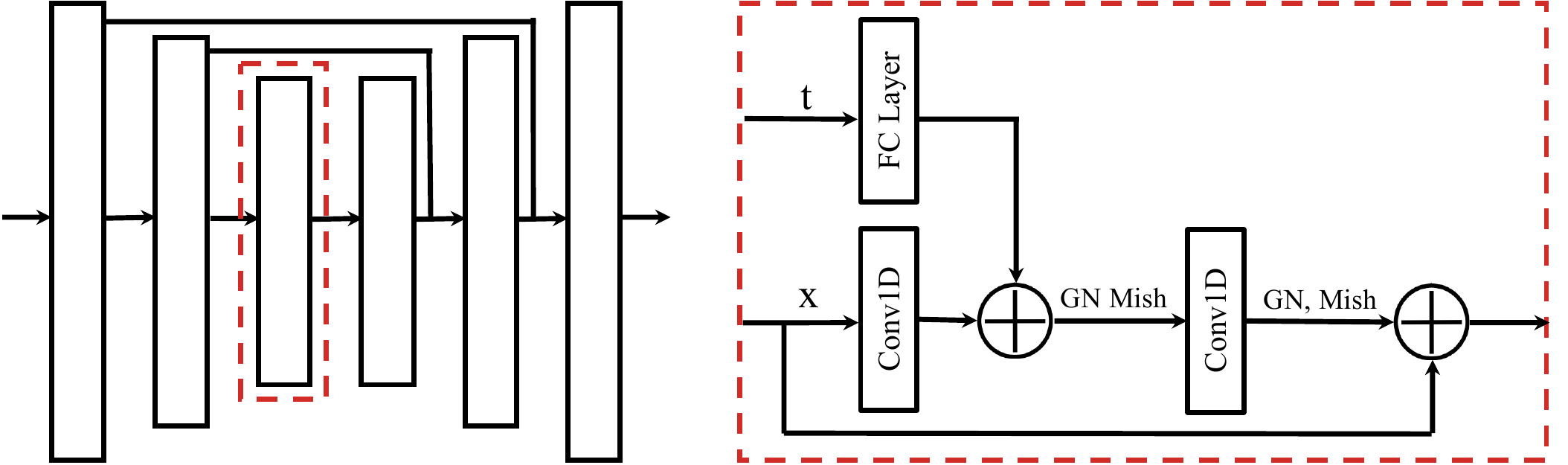}
\caption{
    \method has a U-Net architecture with residual blocks consisting of temporal convolutions, group normalization, and Mish nonlinearities.
}
\label{fig:app_architecture}
\end{figure}

\section{Test-time Flexibility}
\label{app:stacking_costs}

To guide \method{} to stack blocks in specified configurations, we used two separate perturbation functions $h(\btau{})$ to specify a given stack of block A on top of block B, which we detail below.

\myparagraph{Final State Matching} To enforce a final state consisting of block A on top of block B, we trained a perturbation function $h_{\text{match}}(\btau{})$ as a per-timestep classifier determining whether a a state $\vs$ exhibits a stack of block A on top of block B. We train the classifier on the demonstration data as the diffusion model.

\myparagraph{Contact Constraint} To guide the Kuka arm to stack block A on top of block B, we construct a perturbation function $h_{\text{contact}}(\btau{}) = \sum_{i=0}^{64} -1 * \|\btau{}_{c_i} - 1\|^2$, where $\btau{}_{c_i}$ corresponds to the underlying dimension in state $\btau{}_{s_i}$ that specifies the presence or absence of contact between the Kuka arm and block A.  We apply the contact constraint between the Kuka arm and block A for the first 64 timesteps in a trajectory, corresponding to initial contact with block A in a plan.


\section{Implementation Details}

In this section we describe the architecture and record hyperparameters.

\begin{enumerate}
    \item The architecture of Diffuser (Figure~\ref{fig:app_architecture}) consists of a U-Net structure with 6 repeated residual blocks. Each block consisted of two temporal convolutions, each followed by group norm \citep{wu2018groupnorm}, and a final Mish nonlinearity \citep{misra2019mish}.
    Timestep embeddings are produced by a single fully-connected layer and added to the activations of the first temporal convolution within each block.
    \item We train the model using the Adam optimizer \citep{ba2015adam} with a learning rate of $4\mathrm{e}{-05}$ and batch size of $32$.
    We train the models for 500k steps.
    \item The return predictor $\mathcal{J}$ has the structure of the first half of the U-Net used for the diffusion model, with a final linear layer to produce a scalar output.
    \item We use a planning horizon $T$ of 32 in all locomotion tasks, $128$ for block-stacking, $128$ in \texttt{Maze2D / Multi2D U-Maze}, 265 in \texttt{Maze2D / Multi2D Medium}, and 384 in \texttt{Maze2D / Multi2D Large}.
    \item We found that we could reduce the planning horizon for many tasks, but that the guide scale would need to be lowered (\emph{e.g.}, to 0.001 for a horizon of $4$ in the \texttt{halfcheetah} tasks) to accommodate.
    The \href{https://github.com/jannerm/diffuser/blob/34d0e93296c6d8649187e6790ee41cf0c59e3631/config/locomotion.py#L163-L178}{configuration file} in the open-source code demonstrates how to run with a modified scale and horizon.
    \item We use $N=20$ diffusion steps for locomotion tasks and $N=100$ for block-stacking.
    \item We use a guide scale of $\alpha=0.1$ for all tasks except \texttt{hopper-medium-expert}, in which we use a smaller scale of $0.0001$.
    \item We used a discount factor of $0.997$ for the return prediction $\mathcal{J}_\phi$, though found that above $\gamma=0.99$ planning was fairly insensitive to changes in discount factor.
    \item We found that control performance was not substantially affected by the choice of predicting noise $\epsilon$ versus uncorrupted data $\btau{0}$ with the diffusion model.
\end{enumerate}

\end{appendices}

\end{document}